\documentclass{article}

\usepackage[eandd,preprint]{neurips_2026} 
\usepackage[utf8]{inputenc}
\usepackage[T1]{fontenc}
\usepackage{hyperref}
\usepackage{url}
\usepackage{booktabs}
\usepackage{amsmath}
\usepackage{amsfonts}
\usepackage{nicefrac}
\usepackage{microtype}
\usepackage{xcolor}
\usepackage{graphicx}
\usepackage{multirow}
\usepackage{array}
\usepackage{caption}
\usepackage{subcaption}
\usepackage{enumitem}
\usepackage{longtable}
\usepackage{listings}
\usepackage{wrapfig}
\usepackage{placeins}

\lstdefinestyle{guideline}{
  basicstyle=\ttfamily\footnotesize,
  breaklines=true,
  breakatwhitespace=true,
  columns=flexible,
  keepspaces=true,
  showstringspaces=false,
  frame=single,
  framesep=4pt,
  xleftmargin=4pt,
  xrightmargin=4pt,
  upquote=true,
  extendedchars=true,
  inputencoding=utf8,
  literate=%
    {`}{{\textasciigrave}}1
    {—}{{--}}1
    {±}{{$\pm$}}1
    {×}{{$\times$}}1
    {≈}{{$\approx$}}1
    {≤}{{$\leq$}}1
    {≥}{{$\geq$}}1
    {↑}{{$\uparrow$}}1
    {→}{{$\rightarrow$}}1
    {↓}{{$\downarrow$}}1
    {─}{{-}}1
    {│}{{|}}1
    {└}{{+}}1
    {├}{{+}}1
}

\title{How Far Are We From True Auto-Research?}

\author{
\textbf{Zhengxin Zhang}\thanks{Equal contribution; Author order determined by coin flip.} \quad
\textbf{Ning Wang}\footnotemark[1] \quad
\textbf{Sainyam Galhotra} \quad
\textbf{Claire Cardie} \\
Cornell University \\
\texttt{\{zz865, nw366\}@cornell.edu}
}

\begin{document}

\maketitle

\begin{abstract}
Recent auto-research systems \emph{can} produce complete papers, but feasibility is not the same as quality, and the field still lacks a systematic study of how good agent-generated papers actually are. We introduce \textbf{ResearchArena}, a minimal scaffold that lets off-the-shelf agents (Claude Code using Opus 4.6, Codex using GPT-5.4, and Kimi Code using K2.5) carry out the full research loop themselves (ideation, experimentation, paper writing, self-refinement) under only lightweight guidance. Across 13 computer science topic seeds and three trials per agent-topic pair, ResearchArena yields 117 agent-generated papers, each evaluated under three complementary lenses: a manuscript-only reviewer (SAR), an artifact-aware peer review (PR) in which agents inspect the workspace alongside the manuscript, and a human-conducted meta-review. Under SAR alone, the picture is optimistic: Claude Code obtains the highest score, outperforms Analemma's FARS, and matches the weighted-average human ICLR 2025 submission, suggesting that minimally scaffolded agents can produce papers that look competitive on manuscript-only review. Manual inspection, however, reveals this picture is overstated: SAR scores are poorly aligned with its actual acceptance decisions and reward plausible framing without verifying experimental substance. Under artifact-aware PR scores drop sharply, and manual auditing identifies experimental rigor as the major bottleneck, decomposing into three failure modes (\emph{fabricated results}, \emph{underpowered experiments}, and \emph{plan/execution mismatch}) that are highly agent-dependent: Codex 5\%/8\% paper-vs-artifact mismatch / fabricated references versus Kimi Code 77\%/72\%, a $\sim$15$\times$ spread that tracks distinct research \emph{personas} the agents develop. None of the 117 agent-generated papers reaches the acceptance bar of a top-tier venue. This suggests that we are still gaped from the true auto-research.
\end{abstract}

\section{Introduction}

Large language models (LLMs) have rapidly evolved from passive text generators into autonomous agents capable of interleaving reasoning with actions~\cite{yao2022react}, invoking external tools~\cite{schick2023toolformer}, browsing the web~\cite{nakano2021webgpt,zhou2024webarena}, writing and executing code in real software environments~\cite{yang2024sweagent,jimenez2024swebench}, and operating over long horizons in open-ended settings~\cite{wang2023voyager,wang2024agentsurvey}. Recent studies~\cite{lu2024aiscientist,aiscientist_v2,analemma_fars,baek2024researchagent,schmidgall2025agentlab,schmidgall2025agentrxiv} have begun chaining these capabilities into end-to-end scientific research pipelines that take a seed topic and produce a complete research artifact. For example, the AI Scientist~\cite{lu2024aiscientist,aiscientist_v2} brainstorms research ideas, writes and runs code, summarizes results, and drafts full manuscripts, with its second version producing the first peer-review-accepted workshop paper authored entirely by an AI system. Analemma's Fully Automated Research System (FARS)~\cite{analemma_fars} pursues a similar full-pipeline objective at substantially larger compute scale. These works demonstrate that agentic systems \emph{can} produce complete papers. However, they primarily establish feasibility rather than quality: we still lack a systematic study of how good agent-generated papers actually are.

To study this question, we build a minimal scaffold for off-the-shelf agents, called \textbf{ResearchArena}, that lets general-purpose agents carry out the full research loop themselves: ideation, experimentation, paper writing, and self-refinement, with only lightweight guidance. Whereas prior benchmarks such as MLR-Bench~\cite{chen2025mlrbench} evaluate open-ended ML research through modular scaffolds with stage-wise and end-to-end evaluation, ResearchArena studies a complementary setting: a single off-the-shelf agent operates autonomously across broader computer science domains, rather than a pipeline of stage-specific components. We evaluate three frontier  agents: Claude Code with Opus~4.6~\cite{anthropic_opus46}, Codex with GPT-5.4~\cite{openai_gpt54}, and Kimi Code with K2.5~\cite{moonshot_k25}. To test these systems across diverse research settings, we select 13 computer science domains, including 5 CPU-only and 8 GPU-intensive fields, and run 3 trials for each agent-domain pair. This yields 117 agent-generated papers, together with their accompanying with experimental artifacts. Every paper is then evaluated under three complementary lenses: an manuscript-only agentic reviewer (SAR)~\cite{stanford_sar}, our artifact-aware peer review (PR) in which  agents inspect the workspace alongside the manuscript, and human inspection.

SAR-only evaluation paints an optimistic picture and serves as a useful manuscript-only calibration lens: Claude Code obtains the highest average score among the three agents, outperforms Analemma's FARS system, and reaches a score comparable to the weighted-average human-authored ICLR 2025 submission. This suggests that minimally scaffolded agents can produce papers that look competitive under manuscript-only review. Manual inspection, however, reveals that this picture is overstated: SAR scores are poorly aligned with actual ICLR acceptance decisions, and SAR rewards plausible-but-non-workable ideas, polished framing, and honest-looking negative results without verifying experimental substance. Our artifact-aware PR and human inspection tell a different story. Under PR, where reviewers see the code and logs alongside the manuscript, scores drop sharply and almost all papers fall below the acceptance threshold. Manual inspection identifies \emph{experimental rigor} as the major bottleneck across all agents, decomposing into three distinct failure modes: \emph{fabricated results} (numbers reported in the paper do not match the underlying outputs), \emph{underpowered experiments} (narrow scope on a single small dataset and a single model), and \emph{plan/execution mismatch} (the experiment does not include all the components from ideation). These modes are agent-dependent: Codex shows mostly underpowered experiments and the fewest integrity issues (results-vs-artifact mismatches and fabricated references in only 5\% / 8\% of papers), Kimi Code combines fabrication and plan/execution mismatch (77\% / 72\%), and Claude Code falls in between (31\% / 36\%); this $\sim$15$\times$ spread tracks the distinct research \emph{personas} the agents develop: Codex as careful empirical scientist, Kimi Code as ambitious system builder, and Claude Code as full-stack researcher (\S\ref{sec:capabilities}).

Taken together, our findings show that despite producing papers that look polished on manuscript-only review, the actual quality of agent-generated research, measured by artifact-aware peer review and manual auditing, remains far below human-authored work, and \textbf{none of the 117 agent-generated papers reaches the acceptance bar of a top-tier venue}. To support the community in tracking progress as models advance, we release the full corpus: 117 papers with their code and logs, 351 PR reviews, 117 SAR scores, human inspection results, and the configurable harness.

\section{Related Work}

\textbf{Auto-research systems.}
A growing body of work~\cite{karpathy_autoresearch,analemma_fars,lu2024aiscientist,aiscientist_v2,baek2024researchagent,schmidgall2025agentlab,schmidgall2025agentrxiv} has demonstrated end-to-end agent-driven research. 
For example, the AI Scientist~\cite{lu2024aiscientist} pioneered the full loop of ideation, experiments, writing, and automated review with a linear multi-agent pipeline, evaluated on three ML subfields: diffusion modeling, transformer-based language modeling, and learning dynamics. Its successor~\cite{aiscientist_v2} replaces the linear loop with agentic tree search, adds VLM feedback for figures and parallel experiment execution, and produced the first peer-review-accepted workshop paper. Analemma's Fully Automated Research System (FARS)~\cite{analemma_fars}, in contrast, is a closed multi-agent pipeline reportedly run at substantial compute scale (\$104{,}000 reported) and produced over 100 agent-generated papers.
Karpathy's Auto-Research~\cite{karpathy_autoresearch} is, in contrast, a minimal single-agent demonstration that iteratively edits a fixed \texttt{train.py} to explore architecture and hyperparameter choices against a held-out validation metric, automating only the coding-and-experimentation stages.
ResearchAgent~\cite{baek2024researchagent} targets only the early stages (problem definition, method proposal, and experiment design), iteratively refined by multiple LLM-based reviewing agents calibrated to human criteria and grounded in an academic citation graph plus a cross-paper concept store, with no code, experiment execution, or paper writing. Agent Laboratory~\cite{schmidgall2025agentlab} structures the full process as three sequential phases driven by specialized LLM agents: literature review, an \texttt{mle-solver} module for experimentation, and a \texttt{paper-solver} module for report writing, with an optional human-in-the-loop co-pilot mode. AgentRxiv~\cite{schmidgall2025agentrxiv} adds a shared preprint server through which multiple agent laboratories upload and retrieve each other's reports across runs, allowing successive runs to build on prior research rather than operating in isolation.

\textbf{Benchmarks for LLM research agents.}
Existing benchmarks~\cite{huang2024mlagentbench,chan2024mlebench,zhang2025mlrcbench,wijk2024rebench,chen2024scienceagentbench,chen2025mlrbench,starace2025paperbench,siegel2024corebench} evaluate language agents on partial slices of the research process, and fall into three groups. \emph{Fixed-task ML engineering benchmarks} score agents on predefined tasks against objective leaderboard metrics: MLAgentBench~\cite{huang2024mlagentbench} (13 tasks from CIFAR-10 to BabyLM and Kaggle challenges), MLE-bench~\cite{chan2024mlebench} (75 Kaggle competitions), MLRC-Bench~\cite{zhang2025mlrcbench} (7 ML research-competition tasks targeting novel-methodology proposal and implementation), and RE-Bench~\cite{wijk2024rebench} (7 open-ended ML R\&D environments pitting agents against human experts under matched time budgets). \emph{Open-ended research-task benchmarks} draw tasks from peer-reviewed publications and require self-contained research artifacts: ScienceAgentBench~\cite{chen2024scienceagentbench} extracts 102 data-driven discovery problems from 44 papers across four disciplines, and MLR-Bench~\cite{chen2025mlrbench} contains 201 open-ended ML research tasks taken from NeurIPS / ICLR / ICML workshops. \emph{Replication benchmarks} ask agents to reach a known target: PaperBench~\cite{starace2025paperbench} evaluates replicating 20 ICML 2024 spotlights from scratch via hierarchically decomposed rubrics, while CORE-Bench~\cite{siegel2024corebench} measures reproduction of computational results from already-published papers (an adjacent, complementary line of work). 


\section{ResearchArena}
\label{sec:benchmark}
In this secton, we first give an overview of ResearchArena \S\ref{subsec:overview}, describe the setup in \S\ref{subsec:setup} and describe the three complementary evaluation lenses: the Stanford Agentic Reviewer (SAR), our artifacts-aware peer review (PR), and a human inspection in \S\ref{subsec:sar}--\ref{subsec:human}.

\subsection{Overview}
\label{subsec:overview}
\begin{figure}[t]
  \centering
  \includegraphics[width=0.9\linewidth]{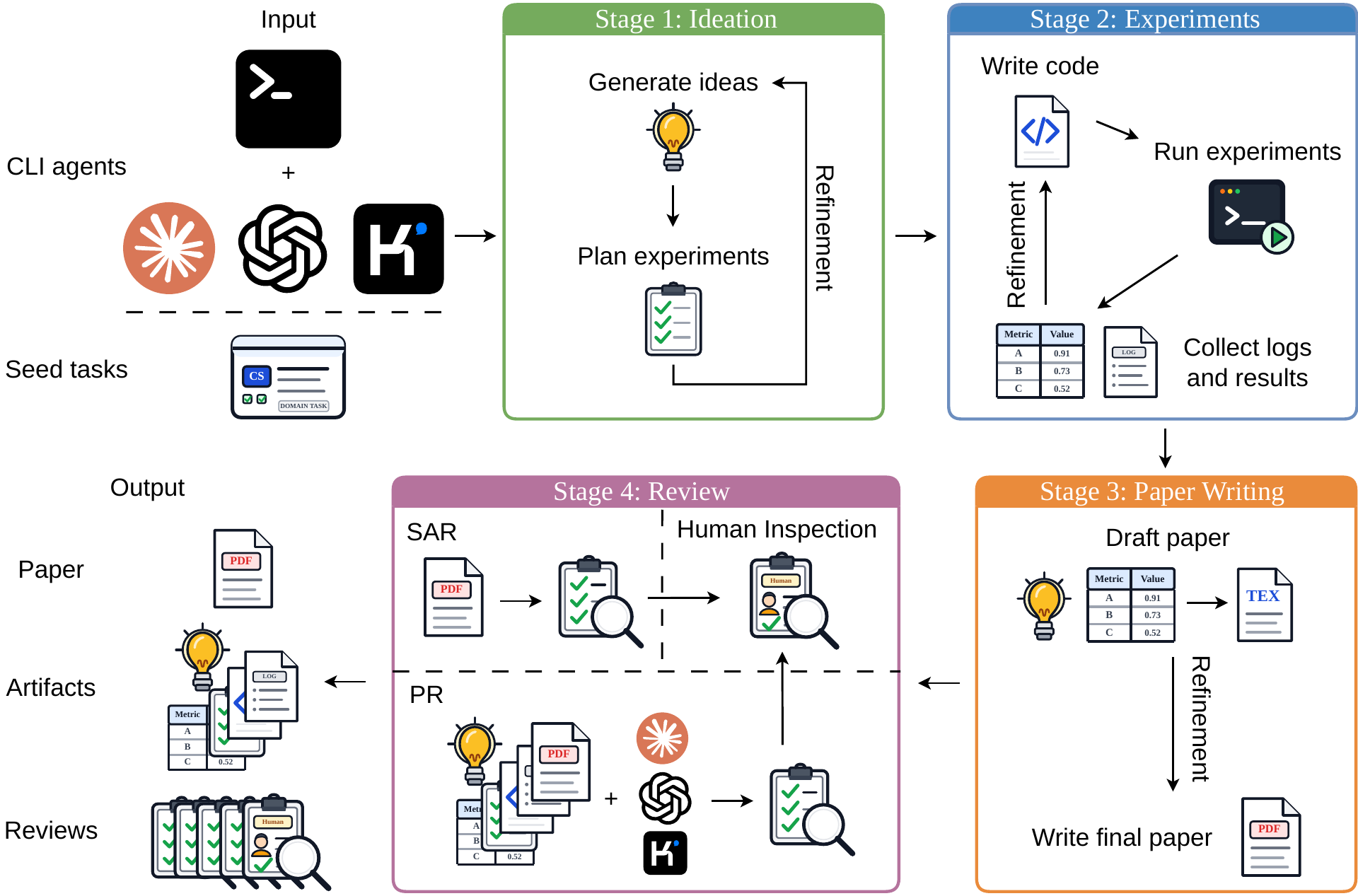}
  \caption{The ResearchArena pipeline.}
  \label{fig:pipeline}
  \vspace{-\baselineskip}
\end{figure}

As shown in Figure~\ref{fig:pipeline}, each  agent receives a CS-domain seed and runs a four-stage research loop: ideation, experiments, paper writing, and review. Stages 1--3 each include a self-refinement loop. At each of these three stages, the agent is paired with a concise domain-specific guideline that fixes the deliverable but not the research itself, distilled from established research practice (e.g., Schulman's ML research notes~\cite{schulman_mlresearch}, the ResearchAgent methodology~\cite{baek2024researchagent}, Peyton Jones's writing advice~\cite{peytonjones_writing}, and the submission and reviewer instructions). The guidelines are intentionally kept short so they act as minimal scaffolding rather than as a step-by-step recipe. We provide example guidelines in Appendix~\ref{app:prompts}. Stage 4 evaluates the resulting paper through three complementary lenses: the Stanford Agentic Reviewer (SAR, manuscript-only), our artifacts-aware peer review (PR, in which three  agents inspect the workspace alongside the manuscript), and human inspection.

\subsection{Setup}
\label{subsec:setup}
ResearchArena spans 13 research seeds across two compute platforms. The 5 CPU seeds (causal learning, compiler optimization, data integration \& cleaning, operating system design, probabilistic methods) target systems / databases / programming-language venues. The 8 GPU seeds (AI for biology, computer vision, datasets \& benchmarks, generative models, interpretability, NLP, privacy in ML, supervised representation learning) target ML venues. Hardware: 1$\times$ NVIDIA RTX A6000 (48~GB) with 4 CPUs and 60~GB RAM for the main experiments. We re-run all GPU seeds on 1$\times$ H100 (80~GB) to test compute scaling (\S\ref{sec:limitations}).

\subsection{Stanford Agentic Reviewer (SAR)}
\label{subsec:sar}
SAR~\cite{stanford_sar} is an automatic agentic paper reviewer that is calibrated to the ICLR scale (0--10) and returns an overall score together with strengthes and weaknesses for any submitted manuscripts. We use SAR for three purposes: (i)~to score all 117 agent-generated papers from the manuscript--only perspective; (ii)~to anchor these scores against human-authored papers by additionally scoring 200 ICLR~2025 papers (100 accepted, 100 rejected); and (iii)~to compare against an existing automated research system by scoring 102 FARS-generated papers.

\subsection{Artifacts-Aware Peer Review (PR)}
\label{subsec:pr}
All three  agents review every paper (351 reviews $= 117 \text{ papers}\times 3$). We distill a domain-specific reviewer guideline (Appendix~\ref{app:reviewer_guideline}), standardize all domains on the ICLR 0--10 scoring scale, and break each review down into nine dimensions: \emph{novelty}, \emph{soundness}, \emph{significance}, \emph{clarity}, \emph{reproducibility}, \emph{experimental rigor}, \emph{references}, \emph{reference integrity}, and \emph{results integrity}. Reviewers check results integrity against experimental artifacts and reference integrity by online lookups against arXiv, Semantic Scholar, and CrossRef. Each reviewer is given \emph{read-only} access to the workspace; the read-only restriction prevents a reviewer agent from silently modifying the artifacts under review, so the paper-vs-artifact comparison reflects what the authoring agent actually produced.

\subsection{Human inspection}
\label{subsec:human}
The authors serve as \emph{meta-reviewers}. For every paper, two authors jointly assess both the manuscript, experimental artifacts, SAR review, and PR reviews. 
The meta-review deliberately focuses on integrity rather than novelty. First, integrity is \emph{objectively verifiable} against the artifacts: a reported number either matches \texttt{results.json} or it does not, and a citation either resolves to a real bibliographic entry or it does not. Novelty, by contrast, are inherently subjective and remain the responsibility of the SAR and PR scores. Second, the official reviewer instructions of top-tier ML conferences all caution reviewers against using ``lack of novelty'' as a sole rejection criterion and ask them to remain open-minded about new ideas; treating novelty as the discriminator for paper quality would therefore run against the field's own reviewing norms.

\section{Capabilities}
\label{sec:capabilities}

In this section, we first compare the three agents against both an automated research system (FARS) and human ICLR papers in \S\ref{subsec:sar_capability}, identify three research personas across the agents in \S\ref{subsec:personas}, and analyze the progamming language and time usage in \S\ref{subsec:pipeline_stats}.

\subsection{Comparison against automated systems and human baselines}
\label{subsec:sar_capability}

\begin{figure}[t]
  \centering
  \begin{minipage}[c]{0.55\linewidth}
    \centering
    \includegraphics[width=\linewidth]{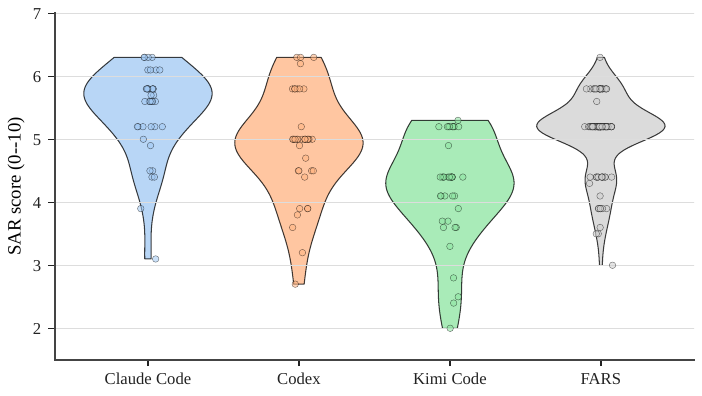}
    \caption{SAR score distributions.}
    \label{fig:sar_violin}
  \end{minipage}\hfill
  \begin{minipage}[c]{0.42\linewidth}
    \captionof{table}{SAR scores.}
    \label{tab:sar}
    \centering
    \small
    \setlength{\tabcolsep}{4pt}
    \begin{tabular}{lccc}
      \toprule
      \textbf{System} & \textbf{n} & \textbf{mean} & \textbf{$\sigma$} \\
      \midrule
      ICLR Accepted                & 100 & \textbf{5.59}          & 0.59 \\
      ICLR Weighted                & 200 & 5.42          & 0.70 \\
      ICLR Rejected                & 100 & 5.34          & 0.75 \\
      \midrule
      \textbf{Claude Code}         &  39 & \textbf{5.45} & 0.70 \\
      FARS~\cite{analemma_fars}    & 102 & 5.06          & 0.62 \\
      Codex                        &  39 & 4.93          & 0.85 \\
      Kimi Code                    &  39 & 4.24          & 0.84 \\
      \bottomrule
    \end{tabular}
  \end{minipage}
\end{figure}
Figure~\ref{fig:sar_violin} shows the SAR score distributions for the three  agents and Analemma's FARS~\cite{analemma_fars}, and Table~\ref{tab:sar} reports the per-system means and standard deviations. Mean scores rank as \textbf{Claude Code (5.45) $>$ FARS (5.06) $>$ Codex (4.93) $>$ Kimi Code (4.24)}: Claude Code outperforms FARS by 0.39 SAR points and Codex achieves similar performance to FARS (4.93 vs.\ 5.06), all while our entire three-agent run cost $\sim$\$1{,}000 ($\approx$\$9 per paper across the 117 papers), versus FARS's reported \$104{,}000 ($\sim$\$1{,}040 per paper), roughly 100$\times$ cheaper per paper. Kimi Code lags the other automated systems. The picture above the ICLR acceptance threshold (SAR $\geq 6$) is even more lopsided: Claude Code produces 21\% (8/39) of papers, versus 10\% for Codex, only 1\% (1/102) for FARS, and 0\% for Kimi Code. Together, these results validate the effectiveness of ResearchArena: a minimal scaffold around an off-the-shelf  agent matches or surpasses a heavily engineered, closed-source auto-research system. Against the 200 ICLR 2025 baselines in Table~\ref{tab:sar}, Claude Code (5.45) sits between rejected (5.34) and accepted (5.59) human submissions and \emph{exceeds} the weighted-average human submission (5.42), where the weighted average mixes the accepted and rejected means in proportion to ICLR's $\sim$32\% acceptance rate. 

\subsection{Three research personas}
\label{subsec:personas}
\begin{figure}[t]
  \centering
  \includegraphics[width=\linewidth]{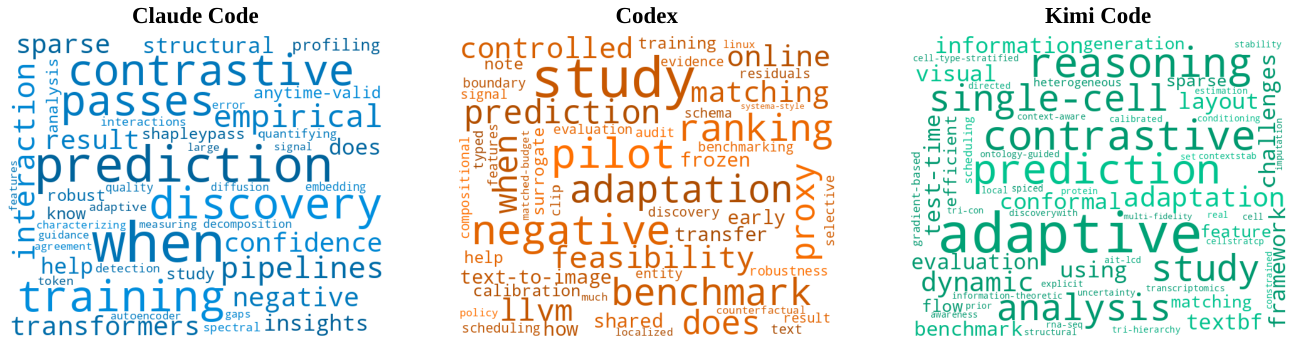}
  \caption{Word cloud of the most frequent content words in each agent's paper titles.}
  \label{fig:persona_wordcloud}
  \vspace{-\baselineskip}
\end{figure}
During our human inspection of all 117 papers (\S\ref{subsec:human}), we find that the three  agents have developed fundamentally different research personas. To make this concrete, we further run research-type analysis on every paper along with the title and the paper structure breakdown, summarized in Table~\ref{tab:personas_summary} and complemented by the per-agent title word cloud in Figure~\ref{fig:persona_wordcloud}.

\begin{table}[!ht]
\centering
\vspace{-\baselineskip}
\caption{Per-agent persona signals on research type, title, and paper structure.}
\label{tab:personas_summary}
\small
\begin{tabular}{lccc}
\toprule
\textbf{Signal}                                 & \textbf{Claude Code} & \textbf{Codex}      & \textbf{Kimi Code}  \\
\midrule
\multicolumn{4}{l}{\emph{Research type}} \\
\quad Method          & 46\%                 & 13\%                & \textbf{79\%}       \\
\quad Benchmark                             &  8\%                 &  0\%                & \textbf{10\%}       \\
\quad Empirical study                           & 46\%                 & \textbf{87\%}       & 10\%                \\
\midrule
\multicolumn{4}{l}{\emph{Title structure}} \\
\quad Avg.\ title length (words)                & 11.3                 & \textbf{11.8}       & 10.2                \\
\quad \% question titles                        & 10\%                 & \textbf{28\%}       &  0\%                \\
\quad \% ``Name: Subtitle'' colon structure     & 74\%                 & 46\%                & \textbf{85\%}       \\
\quad \% acronym-led titles                     & 15\%                 & 49\%                & \textbf{51\%}       \\
\quad \% named-method titles                    & 21\%                 & 38\%                & \textbf{46\%}       \\
\midrule
\multicolumn{4}{l}{\emph{Paper structure}} \\
\quad Paper length (words)                      & \textbf{4{,}023}     & 3{,}421             & 2{,}461             \\
\quad Method-section length (words)             & \textbf{572}         & 531                 & 394                 \\
\quad Equations                                 & 3.8                  & 2.3                 & \textbf{4.0}        \\
\quad Figures                                   & \textbf{4.8}         & 4.1                 & 0.8                 \\
\quad Tables                                    & \textbf{6.0}         & 4.2                 & 4.0                 \\
\quad Algorithm blocks                          & \textbf{0.6}         & 0.0                 & \textbf{0.6}        \\
\quad Theorems / proofs                         & 0.3                  & 0.0                 & \textbf{0.4}        \\
\quad \% with complexity analysis               & \textbf{77\%}        & 10\%                & 64\%                \\
\bottomrule
\end{tabular}
\vspace{-\baselineskip}
\end{table}

\textbf{Claude Code: the full-stack researcher.}
Claude Code produces the most balanced portfolio: 46\% method papers, 46\% empirical studies, and 8\% benchmark papers. It writes the longest papers (4{,}023 words on average) with the most figures (4.8) and tables (6.0), and includes complexity analysis in 77\% of papers. Title style favors essayistic ``\emph{The X of Y}'' framing, e.g.\ \emph{``The Algebra of Compiler Passes: An Empirical Study of Idempotency,''} \emph{``The Bandwidth Knapsack: Optimal Migration Scheduling,''} and \emph{``The Functional Anatomy of Sparse Features in Language Models.''} Title vocabulary leans analytical and mechanistic (\texttt{learning}, \texttt{when}, \texttt{causal}, \texttt{adaptive}, \texttt{pipelines}, \texttt{contrastive}). The full-stack persona is the most ambitious of the three; when Claude Code does fail, the failure mode is narrow-but-occasionally-fabricated experiments rather than wholesale fabrication or method/implementation mismatch (\S\ref{subsec:pr_drop}).

\textbf{Codex: the empirical scientist.}
Codex is overwhelmingly empirical (87\% of papers), while producing only 13\% method papers and \emph{zero} benchmark papers. Its papers are mid-length (3{,}421 words), with the fewest equations (2.3 vs.\ 3.8 / 4.0 for Claude Code / Kimi Code), \emph{zero} algorithm blocks, and \emph{zero} theorems, consistent with an empiricist style that defers from formal claims. Codex has the highest question-title rate at 28\% (vs.\ Claude Code 10\% and Kimi Code 0\%), framed as controlled studies: \emph{``Do Shared Decoders Improve Prototype-Edit Reusability?''}, \emph{``When Does Clarification Supervision Transfer to Formal Reasoning?''}, \emph{``How Much Signal Is in Early Training Trajectories?''}. Title vocabulary clusters around controlled-study and pilot-study terms (\texttt{study}, \texttt{benchmark}, \texttt{negative}, \texttt{matched}, \texttt{controlled}, \texttt{pilot}). The empiricist persona buys high integrity (Codex has the fewest fabricated references; \S\ref{subsec:pr_drop}) but at the cost of empirical breadth: many Codex papers are explicitly scoped as pilot or feasibility studies that are underpowered.

\textbf{Kimi Code: the system builder.}
Kimi Code reframes 79\% of its papers as methods, the highest method-paper rate of any agent. Titles are acronym-heavy named frameworks (51\% acronym rate, 85\% ``Name: Subtitle'' colon structure) and \emph{never} questions: e.g.\ \emph{``CAGER: Causal Geometric Explanation Recovery,''} \emph{``DU-VPT: Decomposed Uncertainty-Guided Visual Prompt Tuning,''} and \emph{``VAST: Velocity-Adaptive Spatially-varying Timesteps.''} Title vocabulary leans toward method-name modifiers (\texttt{adaptive}, \texttt{aware}, \texttt{guided}, \texttt{dynamic}, \texttt{gradient}). Despite the system-builder framing, Kimi Code writes the shortest papers (2{,}461 words) with by far the fewest figures (0.8 on average, sometimes none, vs.\ Claude Code's 4.8) and substitutes formal cues (the most equations at 4.0 and the most theorems at 0.4) for visual evidence. 



\subsection{Programming language and time usage}
\label{subsec:pipeline_stats}

\begin{figure}[h]
  \centering
  \begin{subfigure}[c]{0.66\textwidth}
    \centering
    \includegraphics[width=\linewidth]{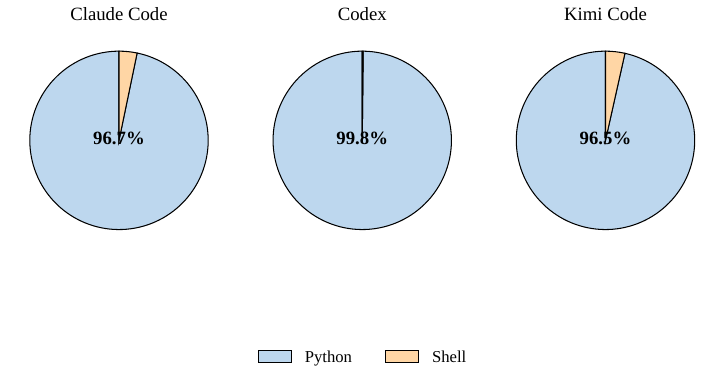}
    \phantomsubcaption
    \label{fig:lang_usage}
  \end{subfigure}\hfill
  \begin{subfigure}[c]{0.33\textwidth}
    \centering
    \includegraphics[width=\linewidth]{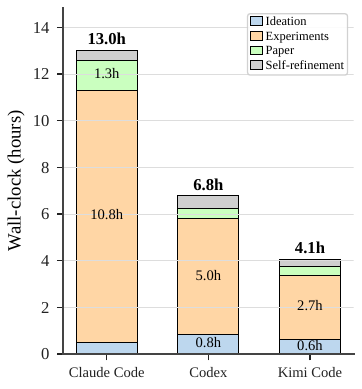}
    \phantomsubcaption
    \label{fig:time_per_stage}
  \end{subfigure}
  \caption{Programming-language usage (left) and wall-clock time per stage (right).}
  \label{fig:pipeline_stats}
\end{figure}

\textbf{Programming-language usage.} 
We further conduct analysis on the programming language of the experiments, shown in Figure~\ref{fig:pipeline_stats} (left). We find that all three agents overwhelmingly default to Python regardless of the research domain with the remainder all shell scripts. Notably, we find zero C/C++/Rust/Go files in any agent's output, even on CPU-only seeds where those languages would be more idiomatic (e.g., C/C++ for operating system design).

\textbf{Wall-clock time per pipeline stage.}
We analyze wall-clock time by pipeline stage for each agent in Figure~\ref{fig:pipeline_stats} (right). All three agents spend the majority of their time on experiments, where Claude Code (13.0h total) is roughly 3$\times$ slower than Kimi Code (4.1h) and 2$\times$ slower than Codex (6.8h). This is consistent with Kimi Code's higher fabrication rate: it does not fully use its compute budget for conducting experiments. Claude Code's longer experimentation time aligns with its lowest underpowered and plan/execution-mismatch rates in \S\ref{subsec:pr_drop}. For ideation, Codex spends the most time; for paper writing, Claude Code takes the longest. Notably, self-refinement takes only a very small share of the total time, almost negligible compared with the other stages.

\section{Limitations}
\label{sec:limitations}

In this section, we first show that SAR alone cannot be trusted as a reliable reviewer (\S\ref{subsec:sar_untrusted}). Artifacts-aware peer reviews and human inspections deliver the three failure modes (\S\ref{subsec:pr_drop}). We then break the scores down by research domain and compute platform (\S\ref{subsec:cpu_gpu_gap}) and rule out compute as the bottleneck (\S\ref{subsec:compute_bottleneck}). A self-refinement and reviewer-severity-drift analysis is in Appendix~\ref{app:selfreview}.

\subsection{SAR cannot be trusted in isolation}
\label{subsec:sar_untrusted}

\begin{wraptable}{r}{0.26\linewidth}
  \centering
  \small
  \vspace{-\baselineskip}
  \setlength{\tabcolsep}{3pt}
  \caption{SAR vs.\ human review.}
  \label{tab:sar_vs_human}
  \begin{tabular}{lcc}
    \toprule
    \textbf{ICLR 2025}           & SAR  & Human \\
    \midrule
    Accepted                     & 5.59 & 6.54 \\
    Rejected                     & 5.34 & 5.02 \\
    $\Delta$ & 0.25 & 1.52 \\
    \bottomrule
  \end{tabular}
  \vspace{-\baselineskip}
\end{wraptable}
We find in Table~\ref{tab:sar_vs_human} that \textbf{SAR is a weaker discriminator than human reviewers}. Comparing SAR scores to the average human review score for each of the same 200 ICLR papers, the human accept-vs-reject score gap is 1.52 points (6.54 vs.\ 5.02), but SAR compresses that gap to only 0.25 points (5.59 vs.\ 5.34). Because SAR does not provide an accept/reject decision, we manually inspect every SAR review and label each paper (Appendix~\ref{app:manual_sar_decisions}). The resulting acceptance rates make the same point: SAR accepts 76\% of human-accepted ICLR papers and 52\% of human-rejected ones, but only 41\% of Claude Code's, 22\% of FARS's, 13\% of Codex's, and 5\% of Kimi Code's. Mean scores overstate how close agents are to top-tier acceptance; the underlying acceptance gap is much larger, and SAR cannot be the sole evaluator of agent-generated papers.

\subsection{Artifacts-aware peer review surfaces three failure modes}
\label{subsec:pr_drop}
\begin{figure}[t]
  \centering
  \vspace{-\baselineskip}
  \begin{minipage}[t]{0.49\linewidth}
    \centering
    \includegraphics[width=\linewidth]{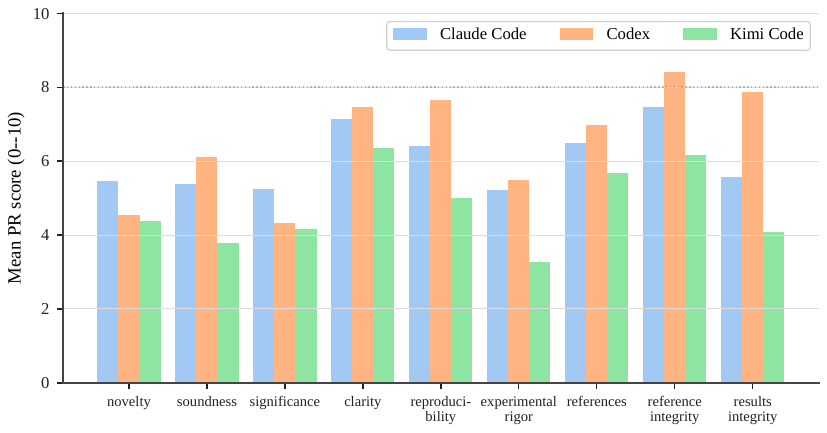}
    \captionof{figure}{PR breakdown scores.}
    \label{fig:per_dimension}
  \end{minipage}\hfill
  \begin{minipage}[t]{0.49\linewidth}
    \centering
    \includegraphics[width=\linewidth]{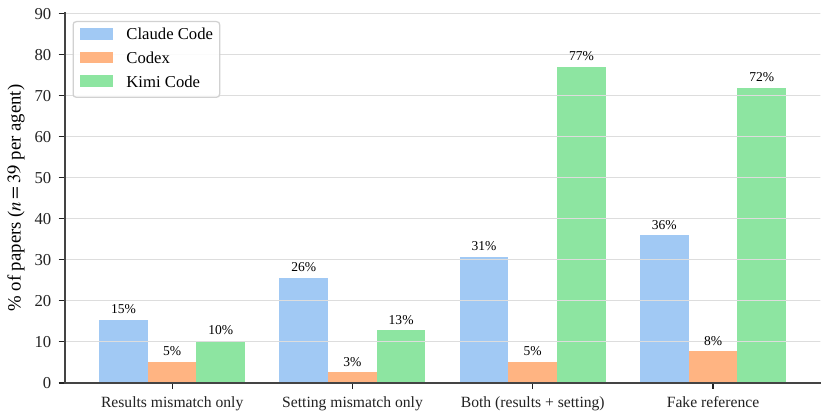}
    \captionof{figure}{Fabricated results.}
    \label{fig:integrity}
  \end{minipage}
\end{figure}
\begin{wrapfigure}{r}{0.42\linewidth}
  \centering
  \vspace{-\baselineskip}
  \includegraphics[width=\linewidth]{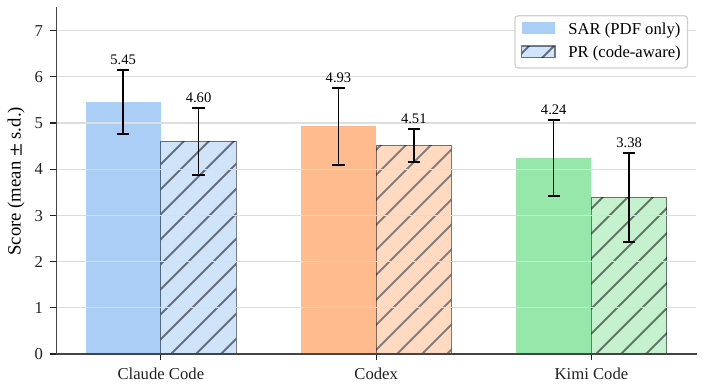}
  \caption{SAR vs.\ PR.}
  \label{fig:sar_vs_pr}
  \vspace{-\baselineskip}
\end{wrapfigure}
Under artifacts-aware PR review, every agent's score drops below its SAR score (Figure~\ref{fig:sar_vs_pr}): Claude Code $-0.85$, Codex $-0.42$, Kimi Code $-0.86$. Through per-dimension PR scores (Figure~\ref{fig:per_dimension}) localise the drop: Codex leads on every reliability-leaning dimension (reproducibility, references, reference and results integrity), Claude Code leads on creative dimensions (novelty, significance), Kimi Code lags on every dimension simultaneously, and \textbf{experimental rigor is the lowest dimension across all agents}. To further investigate the experiment rigor problems. 
We manually verify three failure modes (\emph{fabricated results}, \emph{underpowered experiments}, and \emph{plan/execution mismatch}).

\textbf{Fabricated results.} We classify fabricated results into 4 categories: \emph{results mismatch only} (numbers reported in the paper do not match \texttt{results.json}, logs, or experiment outputs), \emph{setting mismatch only} (the paper claims components not implemented in the code, or hyperparameters in the text differ from the config), \emph{both} (the paper exhibits both results and setting mismatches), and \emph{fake reference} (citations that do not exist, have fabricated authors, or have incorrect bibliographic metadata). As shown in Figure~\ref{fig:integrity}, Kimi Code shows by far the highest rates (77\% paper-vs-artifact mismatch, 72\% fake references): it invents experimental results directly (\hyperref[case:4]{Case~4}) or reports baselines that were never run (\hyperref[case:5]{Case~5}). Claude Code follows at 31\%/36\% (occasional fabrication when experiments fail; \hyperref[case:2]{Case~2}); Codex stays clean at 5\%/8\%.

\textbf{Underpowered experiments.} A paper is flagged as \emph{underpowered} by having limited experiments. For example, a single small dataset where multiple are expected, one model size where a ladder is expected, or one random seed for what should be a stochastic comparison, or when the paper is explicitly framed as a pilot or feasibility study with limited evidence. As shown in Table~\ref{tab:failure_modes}, Kimi Code 82.1\% $>$ Codex 41.0\% $>$ Claude Code 25.6\%; for every agent the rate is higher on GPU than on CPU (Claude Code 33\% vs.\ 13\%, Codex 42\% vs.\ 40\%, Kimi Code 92\% vs.\ 67\%), consistent with GPU work being broader in scope than the CPU-only seeds. Codex's papers are often explicitly framed as pilot/feasibility studies (\hyperref[case:3]{Case~3}), which reduces fabrication but limits the empirical evidence.

\textbf{Plan/execution mismatch.} During ideation, each agent is aware of the resources (i.e., hardware and time budget) to write an experimental plan based on the proposed ideas. We define \emph{plan/execution mismatch} as cases where the executed artifacts diverge from that plan or from the manuscript that follows it: a baseline named in the plan but missing from the code, an ablation specified but never run. As shown in Table~\ref{tab:failure_modes}, Kimi Code 33.3\% $>$ Codex 20.5\% $>$ Claude Code 17.9\%; the CPU-vs-GPU pattern differs by agent (Claude Code 13\% / 21\%, Codex 33\% / 13\%, Kimi Code 33\% / 33\%): Claude Code's mismatch concentrates on GPU work, Codex's on CPU, while Kimi Code splits evenly.
Kimi Code plans the most experiments (13.2/trial), and its overambitious planning exceeds what it can execute, producing the highest plan/execution mismatch (33.3\%) and underpowered rate (82.1\%). Codex plans the most conservatively (5.7/trial), which keeps mismatch low (20.5\%) but inflates the underpowered rate (41.0\%) via frequent pilot/feasibility framing. Claude Code plans moderately (10.6/trial) and has the lowest rates on both axes (17.9\% mismatch, 25.6\% underpowered).
A shared failure across all agents is the tendency to compare against \emph{older} baselines rather than recent ones, even when newer baselines are mentioned in the related-work section.

Besides the three failure modes, we observe that PR review and SAR review both credit the honesty towards negative results, where human reviewers would not credit this as a strength.


\begin{table}[t]
\centering
\caption{Per-agent breakdown of \emph{underpowered} and \emph{plan/execution mismatch} ratios.}
\label{tab:failure_modes}
\begin{tabular}{lccccccccc}
\toprule
            &     & \multicolumn{3}{c}{\textbf{Underpowered (\%)}} & \multicolumn{3}{c}{\textbf{Plan/exec.\ mismatch (\%)}} & \textbf{Avg planned} \\
\cmidrule(lr){3-5} \cmidrule(lr){6-8}
\textbf{Agent} & $\boldsymbol{n}$ & \textbf{CPU} & \textbf{GPU} & \textbf{Total} & \textbf{CPU} & \textbf{GPU} & \textbf{Total} & \textbf{exp./trial} \\
\midrule
Claude Code & 39  & 13.3 & 33.3 & 25.6  & 13.3 & 20.8 & 17.9  & 10.6 \\
Codex       & 39  & 40.0 & 41.7 & 41.0  & 33.3 & 12.5 & 20.5  &  5.7 \\
Kimi Code   & 39  & 66.7 & 91.7 & 82.1  & 33.3 & 33.3 & 33.3  & 13.2 \\
\midrule
Overall     & 117 & 40.0 & 55.6 & 49.6  & 26.7 & 22.2 & 23.9  &  9.8 \\
\bottomrule
\end{tabular}

\end{table}

\subsection{Per-domain analysis}
\label{subsec:cpu_gpu_gap}

\begin{figure}[h]
  \centering
  \includegraphics[width=0.9\linewidth]{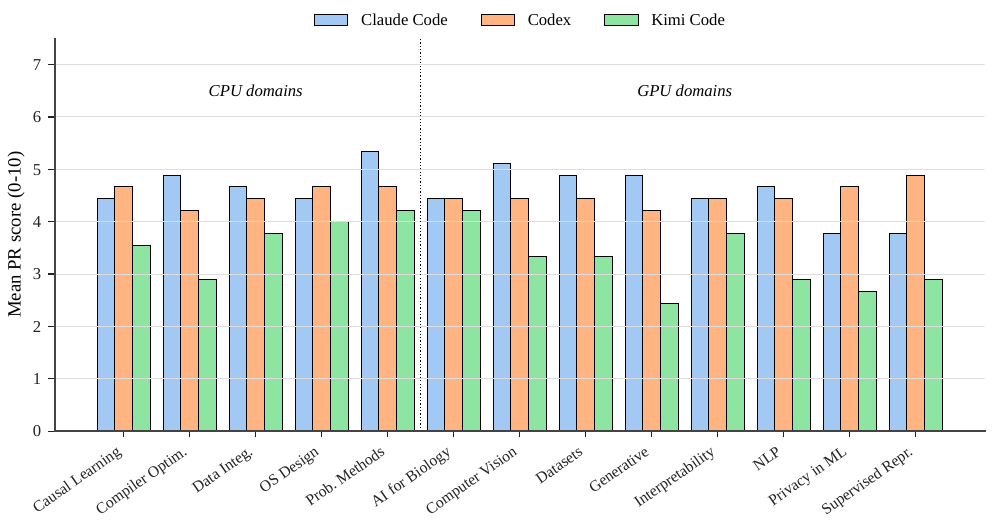}
  \caption{Per-domain mean PR scores by agent across the 13 research domains.}
  \label{fig:per_domain_breakdown}
  \vspace{-\baselineskip}
\end{figure}

\textbf{Per-domain breakdown.} Figure~\ref{fig:per_domain_breakdown} shows mean PR scores per agent across the 13 research domains (the parallel SAR breakdown is in Appendix~\ref{app:per_domain} due to the space limit). Patterns vary by agent. Claude Code peaks on Probabilistic Methods (5.32) and Computer Vision (5.10) but dips to $\sim$3.78 on Privacy in ML and Supervised Repr.\ Learning. Codex stays in a tighter band (4.20--4.89), with its highest mark on Supervised Repr.\ Learning (4.89). Kimi Code is consistently the lowest, with its weakest scores on Generative Models (2.44) and Privacy in ML (2.66). 

\textbf{CPU vs.\ GPU: opposite trends in SAR and PR.} \textbf{PR and SAR move in opposite directions across the CPU/GPU split} (Table~\ref{tab:cpu_gpu} in Appendix~\ref{app:per_domain}). Under \emph{PR}, all three agents score higher on CPU than on GPU (Claude Code $+0.26$, Codex $+0.03$, Kimi Code $+0.50$), with Kimi Code showing the largest gap. Under \emph{SAR}, Codex and Kimi Code score \emph{higher} on GPU (Codex $-0.61$, Kimi Code $-0.20$), while only Claude Code is roughly platform-invariant. GPU domains (vision, NLP, generative models) are well-established fields where agents can produce better-looking papers (more polished prose, more figures, familiar baselines), but GPU experiments are also harder to execute correctly: CUDA issues, memory limits, and training instabilities lead to more incomplete runs and mismatched results when reviewers verify the code. CPU tasks are simpler to run and verify, yielding more reliable experiments. This divergence further illustrates that SAR alone is insufficient: it rewards presentation quality over experimental substance.

\subsection{Compute is not the bottleneck}
\label{subsec:compute_bottleneck}
We re-run all 8 GPU seeds with Codex on 8$\times$ NVIDIA H100 (80~GB) for 3 trials each, with budget matched to the A6000 runs. The result is no consistent improvement: Codex PR drops from 4.51 (A6000) to 4.26 (H100), confirming that the limiting factor is not compute but the agent's experiment design capabilities. The per-domain H100 vs.\ A6000 breakdown is in Appendix~\ref{app:h100}.

\section{Future Directions}
\label{sec:future}

\textbf{Can we trust agentic reviewers for agent-generated papers?} SAR and PR both over-credit agent-generated papers relative to human reviewers (\S\ref{subsec:sar_untrusted}). Future automated reviewers should combine with principled calibration against human review.

\textbf{Faithfulness over complex tasks.} Frontier model providers increasingly advertise faithfulness as a core capability of their agents, yet under our open-ended end-to-end research setting we still observe substantial fabrication (\S\ref{subsec:pr_drop}). The claim of faithful behaviour does not yet survive contact with sufficiently complex tasks. Future work should focus on training agents to be faithful end-to-end rather than only on individual reasoning traces. 

\textbf{Better experiment-planning agents and scaffolds.} The major challenge for auto-research is experimental rigor (\S\ref{subsec:pr_drop}). Closing this gap will require improvements: stronger agent capabilities and scaffolds that harness agents for designing and executing rigorous experiments end-to-end.








\section{Conclusion}
In this paper, we systematically investigate the auto-research capabilities and limitations of three frontier  agents across 13 CS domains using a minimal scaffold, ResearchArena. We find that experimental rigor is the number-one weakness: agents routinely fail to plan, execute, and faithfully report experiments, limiting both the scope and significance of their papers. Fabricated results and a manuscript-only reviewer that systematically favours honest but narrow framings further raise faithfulness concerns for today's frontier models. In terms of paper quality, all current agents still fall well short of the threshold for top-tier venues. There is still a long way to go for true auto-research.

\bibliographystyle{plain}
\bibliography{references}

\clearpage
\appendix

\section{Limitations}
\label{app:limit}
Due to budget constraints, our study evaluates only three  agents and therefore does not cover the full space of available agentic coding systems. We also focus on computer-science research domains, leaving evaluation in other scientific and engineering fields to future work. Finally, our end-to-end analysis is conducted under our specific ResearchArena setting; although this setting reflects many emerging auto-research workflows, the findings may not fully generalize to all possible auto-research systems, scaffolds, or evaluation protocols.

\section{Per-stage guidelines}
\label{app:prompts}

Each pipeline stage is anchored by a domain-aware guideline document. We reproduce the ML-domain guidelines below verbatim as the canonical reference; the other five domain families (systems, databases, PL, theory, security) follow the same structure with domain-specific phrasing.

\paragraph{Ideation guidelines (stage 1).} The agent proposes a research idea framed as a hypothesis with a falsifiable prediction; output is \texttt{idea.json}. Distilled from Schulman's research advice~\cite{schulman_mlresearch}, the human study of LLM ideation by Si et al.~\cite{si2024llmideas}, and the ResearchAgent methodology~\cite{baek2024researchagent}.

\begin{lstlisting}[style=guideline]
# Idea Generation Guidelines

How to go from a seed field to a novel, feasible research idea.

Distilled from John Schulman's "Opinionated Guide to ML Research", the
ResearchAgent methodology, "Can LLMs Generate Novel Research Ideas?" (Si et al.),
and standard academic research practices.

## Step 1: Explore the field

Start by understanding what already exists. DO NOT skip this step.

### Search for existing work (newest to oldest)
- Search arXiv (arxiv.org), Semantic Scholar (semanticscholar.org), and
  Google Scholar (scholar.google.com) for papers in your seed field
- **Start with the newest papers first** — sort by date, read the most
  recent work before going to older foundational papers. This ensures
  you know the current frontier before proposing something new.
- Recommended search order:
  1. Last 6 months — what's happening right now?
  2. Last 1-2 years — what are the current state-of-the-art methods?
  3. Foundational papers — what are the classic approaches?
- Look for:
  - Survey papers — they summarize the landscape and list open problems
  - Highly-cited recent papers — they define the current state of the art
  - Workshop papers — they often contain early-stage ideas and emerging trends

### Build a mental map
- What are the main approaches in this area?
- What are the established benchmarks and metrics?
- What are the known limitations of current methods?
- What problems are considered "open" or "unsolved"?
- What recent techniques from OTHER fields could apply here?

### Find the gaps
- Read the "Limitations" and "Future Work" sections of recent papers
- Look for recurring complaints in reviews (on OpenReview, if available)
- Identify assumptions that current methods make — can you relax them?
- Look for problems where simple baselines still perform surprisingly well
  (this signals the community hasn't cracked it yet)

## Step 2: Generate candidate ideas

### Two approaches (choose one or combine)

**Goal-driven** (recommended): Start with a problem you want to solve.
- "Current methods for X fail when Y happens. How can we fix that?"
- "Task Z requires too much labeled data. Can we do it with less?"
- The goal constrains your search and makes the contribution clear.

**Idea-driven**: Start with a technique and find where it applies.
- "Technique A works well for B. Could it also work for C?"
- Riskier — you may find the idea already exists or doesn't work.

### What makes a good research idea
- **Novel**: Not already done. You MUST verify this (Step 3).
- **Feasible**: Can be implemented and tested within your resource constraints.
- **Clear**: The contribution is easy to explain in one sentence.
- **Testable**: There's a concrete way to evaluate whether it works.
- **Significant**: If it works, the community would care.

### What makes a BAD research idea
- Too broad ("improve NLP") — needs a specific problem and approach
- Too incremental ("change hyperparameter X from 0.1 to 0.01")
- Not verifiable (no way to test if it worked)
- Requires resources you don't have (100 GPUs, proprietary data)
- Already exists (you didn't check the literature)

## Step 3: Verify novelty (CRITICAL — do not skip)

Before committing to an idea, verify it hasn't been done:

### Search specifically for your idea
- Search Semantic Scholar and arXiv with keywords from your proposed method
- Search for the PROBLEM you're solving, not just your approach
- Check if your idea is a special case of something more general that exists
- Look at the "Related Work" sections of papers closest to your idea

### Common novelty traps
- Your idea exists but under a different name (jargon varies across subfields)
- Your idea was tried and didn't work (check for negative results too)
- Your idea is a minor variation of an existing approach
- A concurrent paper (posted in the last few months) does the same thing

### If your idea already exists
- DON'T give up immediately. Ask: can you improve on it? Apply it to a
  new domain? Combine it with something else? Scale it up?
- If it truly exists with no room for improvement, go back to Step 2

## Step 4: Produce outputs

You must produce THREE outputs in this step:

### 4.1 proposal.md — Research Proposal

A thorough document with these sections:
- **Introduction**: Context, problem statement, key insight, hypothesis
- **Proposed Approach**: Overview, method details, key innovations
- **Related Work**: Key papers, how your idea differs, positioning
- **Experiments**: Planned setup, benchmarks, metrics, expected results
- **Success Criteria**: What would confirm or refute your hypothesis
- **References**: Full citation list (all must be real, verifiable papers)

### 4.2 idea.json — Structured Summary

A JSON object with at least these fields:
- **description**: 1-3 sentences explaining what you're proposing
- **title**: paper title
- **motivation**: why this problem matters, what gap you're filling
- **proposed_approach**: your high-level method and why it should work
- **related_work**: key existing papers and how your idea differs
  (use REAL papers you found in Steps 1 and 3 — include titles and authors)
- **hypothesis**: testable hypothesis
- **success_criteria**: what would confirm/refute the hypothesis

### 4.3 references/ — Parsed Reference Papers

Create a directory with key reference papers. For each paper, create a
subdirectory containing the paper's content parsed into sections:
```
references/
├── Paper-Title-One/
│   ├── meta/
│   │   ├── meta_info.txt       # title, authors, venue, year, URL
│   │   └── bibtex.txt          # BibTeX entry
│   └── sections/
│       ├── abstract.md
│       ├── 1 Introduction.md
│       ├── 2 Related Work.md
│       └── ...
├── Paper-Title-Two/
│   └── ...
```

This grounds your proposal in real literature and ensures references are
verifiable by reviewers.

### Sanity checks before moving on
- Can you explain the idea in one sentence to a non-expert?
- Is there a clear experiment that would test the idea?
- Do you have the resources (data, compute, time) to do it?
- Is the expected contribution large enough for a paper?
- Are all references real, verifiable publications?

## General principles

### From John Schulman
- Your ability to choose the right problem is more important than raw skill
- Watch which ideas prosper and which are forgotten — this develops taste
- Goal-driven research has lower scooping risk than idea-driven research
- There's no shame in working on ideas suggested by others or by the literature

### From "Can LLMs Generate Novel Research Ideas?" (Si et al.)
- AI-generated ideas tend to be novel but lack feasibility — ground yours
  in practical constraints
- Vague implementation details are the #1 weakness — be specific about how
  your method actually works
- Missing baselines and unrealistic assumptions are common failures
- Verify your idea against existing work — 80% of reviewer rejections
  cite existing papers that the authors missed

### From ResearchAgent (Baek et al.)
- Connect ideas across papers, not just within one paper
- Look for shared concepts across different subfields
- Iterative refinement improves idea quality — but diminishing returns
  after 2-3 rounds
- Both citation relationships AND underlying concepts matter for novelty
\end{lstlisting}

\paragraph{Plan guidelines (stage 2, planning).} The agent converts the idea into a concrete experiment plan covering datasets, baselines, metrics, expected outcomes under the hypothesis, ablations, and statistical tests; output is \texttt{plan.json}.

\begin{lstlisting}[style=guideline]
# Experiment Plan Guidelines

How to design a rigorous experiment plan based on your research proposal.

Read proposal.md and idea.json first — your plan must test the claims
and hypothesis described there.

## Plan Format

Save your plan as plan.json — a JSON array of experiment steps:
```json
[
  {
    "category": "<category>",
    "title": "short descriptive title",
    "description": "what this step does and why",
    "steps": {
      "step1": "detailed instruction with specifics",
      "step2": "...",
      ...
    }
  },
  ...
]
```

Suggested categories (add your own as needed):
- **Environment Configuration** — dependencies, setup
- **Data Preparation** — download, preprocess, splits, statistics
- **Baseline Experiment** — existing methods to compare against
- **Main Experiment** — your proposed method
- **Analysis Experiment** — ablations, robustness, sensitivity
- **Effectiveness Evaluation** — success criteria, statistical tests
- **Visualization** — figures, tables, plots for the paper

## Experiment Design Principles

### Formulate testable claims
- State your hypothesis as a testable claim
- Design experiments that could fail — if they can't produce a negative
  result, they're not informative
- Define what would DISPROVE your claim

### Choose the right experiment type

| Claim type | Experiment type | What to measure |
|---|---|---|
| "Our method outperforms X" | Empirical comparison | Metrics on shared benchmarks |
| "Component A is critical" | Ablation study | Performance with/without A |
| "This scales better" | Scaling experiment | Performance vs. data/compute/params |
| "Our theory predicts X" | Theoretical validation | Synthetic setup with known ground truth |
| "This property holds" | Analysis/probing | Measurements on existing models/data |
| "This is faster/cheaper" | Systems experiment | Latency, throughput, memory, FLOPs |
| "This benchmark is better" | Benchmark evaluation | Existing methods on new benchmark |

### Select metrics carefully
- Use standard metrics for your task
- Report ALL standard metrics, not just the one where you win
- Consider both performance AND cost (FLOPs, latency, memory)

### Choose datasets that test your claim
- Use standard benchmarks when possible
- If claiming robustness, test on distribution-shifted data
- If claiming generalization, test on multiple datasets
- Document: data source, size, splits, preprocessing

### Select baselines fairly
- At least 2 meaningful baselines (one simple, one strong/recent)
- Run all baselines with equivalent effort (same compute, same tuning)
- Never compare against intentionally weak baselines

### Plan ablation studies
- For each novel component, plan to remove it and measure impact
- Plan ablations BEFORE running experiments, not after seeing results

### Think about confounders
- Could the improvement come from more parameters/data/compute?
- Are comparisons fair (same preprocessing, splits, compute budget)?
- If using published baselines, are the setups truly comparable?

### Rigorous evaluation
- Use a fixed random seed for reproducibility
- Use the SAME seed for method and all baselines (fair comparison)
- Avoid data leakage (preprocessing stats from train only, etc.)
- Test set used ONCE for final evaluation, not for model selection

### Common pitfalls to avoid
- Don't tune hyperparameters on the test set
- Don't compare against baselines with different preprocessing or splits
- Don't report only the metric where you win
- Don't claim SOTA without comparing against actual SOTA methods
- Don't ignore negative results — report them honestly
- Don't use a single train/test split
- Don't assume deep learning is always better — test simpler alternatives

## Feasibility and Runtime Estimation

Before finalizing the plan, estimate whether it fits the resource budget:

1. **Estimate per-experiment runtime**: How long does one training run /
   evaluation / API call take? Use published benchmarks or rough estimates
   based on model size and dataset size.
2. **Count independent experiments**: How many baselines, seeds, ablations,
   and datasets? Identify which can run in parallel.
3. **Divide by available parallelism**: If you have K GPUs and N independent
   GPU experiments, parallel runtime ≈ N/K × per-experiment time. For CPU
   tasks, divide by available cores.
4. **Compare to budget**: If estimated parallel runtime exceeds the time
   budget, simplify the plan (fewer seeds, smaller models, fewer datasets)
   BEFORE finalizing — not during execution.

Example:
- 5 baselines + 1 method + 3 ablations = 9 experiments
- Each run: ~30 min on 1 GPU
- With 8 GPUs: 9/8 ≈ 2 batches × 30 min = ~1 hour
- Budget is 8 hours → plenty of room for visualization + analysis

## Plan Quality Checklist

Before finalizing plan.json, verify:
- [ ] Each step has a clear category, title, description, and sub-steps
- [ ] Sub-steps are detailed enough to follow without ambiguity
- [ ] Specific datasets, metrics, and hyperparameters are named
- [ ] At least 2 meaningful baselines are included
- [ ] Ablation studies planned for each novel component
- [ ] Fixed random seed for reproducibility
- [ ] Success criteria clearly defined
- [ ] Runtime estimate fits the resource budget with parallel execution
- [ ] Plan accounts for ALL available GPUs and CPUs
- [ ] Visualization step included for paper figures
\end{lstlisting}

\paragraph{Experiment guidelines (stage 2, execution).} The agent writes self-contained Python under \texttt{exp/} that prepares data, trains baselines and the proposed method, runs ablations, and writes \texttt{results.json} containing all reported metrics; per-command logs are kept.

\begin{lstlisting}[style=guideline]
# Experiment Execution Guidelines

How to execute your experiment plan efficiently and rigorously.
Experiment design principles are in plan_guidelines.md — you should
have already applied them when creating plan.json.

## Phase 1: Maximize Resource Usage

Your goal is to use ALL available resources efficiently. Check what you
have before starting:
```bash
nvidia-smi          # GPUs: count, memory, current usage
nproc               # CPU cores
free -h             # RAM
```

### Parallel execution strategy

Identify independent experiments in your plan (different seeds, baselines,
ablations, datasets) and run them simultaneously:

```bash
# Example: 8 GPUs, 6 independent experiments
CUDA_VISIBLE_DEVICES=0 python exp/baseline1/run.py &
CUDA_VISIBLE_DEVICES=1 python exp/baseline2/run.py &
CUDA_VISIBLE_DEVICES=2 python exp/method/run.py &
CUDA_VISIBLE_DEVICES=3 python exp/ablation1/run.py &
CUDA_VISIBLE_DEVICES=4 python exp/ablation2/run.py &
CUDA_VISIBLE_DEVICES=5 python exp/dataset2/run.py &
wait  # wait for all to finish
```

For CPU-bound work (data preprocessing, evaluation, API calls):
```bash
# Run multiple CPU tasks in parallel
python exp/preprocess_dataset1.py &
python exp/preprocess_dataset2.py &
python exp/preprocess_dataset3.py &
wait
```

### GPU utilization
- **Pin experiments to GPUs** with `CUDA_VISIBLE_DEVICES=N`
- If a model uses only part of GPU memory, run multiple experiments
  per GPU (e.g., 2 small models on one 48GB GPU)
- Increase batch size to fill GPU memory — larger batches = faster training
- For inference-only experiments (embedding, evaluation), consider
  running several on the same GPU

### CPU utilization
- Use `multiprocessing` or `concurrent.futures.ProcessPoolExecutor` for
  CPU-bound data processing
- Parallelize data loading with `num_workers` in PyTorch DataLoaders
- For API-based experiments (LLM scoring), use `asyncio` or thread pools
  to make concurrent API calls

### Follow the plan — do not scope down
Your plan.json was designed with the available resources in mind.
Execute ALL steps. If a step truly cannot run (dependency failure,
out-of-memory), document it in that step's `SKIPPED.md` and move on.
Do NOT drop experiments just because a single-GPU pilot looks slow —
use parallel execution across all GPUs.

### Prioritize execution order
Run experiments in dependency order:
1. Data preparation (must finish first)
2. Baselines + method (can run in parallel)
3. Ablations (can run in parallel after method works)
4. Analysis + visualization (after results are in)

## Phase 2: Workspace Structure

Organize experiments so that each step has its own folder with code, results,
and logs. This makes it easy to verify which code produced which results.

```
exp/
├── <experiment_name>/              # one folder per experiment/condition
│   ├── run.py                      # experiment script
│   ├── config.yaml (or .json)      # hyperparameters, settings
│   ├── results.json                # per-experiment results
│   └── logs/                       # training/eval logs, stdout
│
├── <baseline_name>/
│   ├── run.py
│   ├── results.json
│   └── logs/
│
├── <ablation_name>/
│   ├── run.py
│   ├── results.json
│   └── logs/
│
└── shared/                         # shared utilities across experiments
    ├── data_loader.py              # data loading, preprocessing
    ├── metrics.py                  # evaluation metrics
    ├── models.py                   # model definitions
    └── utils.py                    # common helpers

data/                               # downloaded/processed datasets
figures/                            # generated figures for the paper
```

### Per-experiment results

Each `exp/<name>/results.json` should capture that experiment's output:
```json
{
  "experiment": "<name>",
  "metrics": {"metric1": {"mean": 0.87, "std": 0.002}, ...},
  "config": {"lr": 0.001, "epochs": 50, "seed": 42, ...},
  "runtime_minutes": 45
}
```

### Figures

Save publication-ready figures to `figures/`:
- Comparison plots (your method vs baselines)
- Ablation charts (impact of each component)
- Training curves (loss/metric over epochs)
- Analysis visualizations (distributions, embeddings, etc.)

Each figure should be self-contained with axis labels, legends, and titles.

## Phase 4: Plan Compliance

Follow plan.json step by step:
- Execute every step in order
- Create a subfolder under `exp/` for each plan step
- If a step is infeasible, document why in that step's folder (create a
  `SKIPPED.md` with the reason) and move on
- After all steps, verify that the plan's success criteria are met
- If results contradict the hypothesis, report this honestly — negative
  results with good analysis are valuable

## Reproducibility Checklist

Before finishing, verify:
- [ ] Fixed random seeds used throughout
- [ ] At least 2 meaningful baselines compared fairly
- [ ] Fixed random seed used for reproducibility
- [ ] Ablation study for each novel component
- [ ] No data leakage (verified)
- [ ] All configuration documented in per-experiment results
- [ ] Each experiment has its own folder under exp/ with code and results
- [ ] Figures saved for key results
- [ ] Negative results reported honestly (if any)
\end{lstlisting}

\paragraph{Paper-writing guidelines (stage 3).} The agent produces a NeurIPS-style \texttt{paper.tex} that cites the items in \texttt{references.bib}; reported numbers must match \texttt{results.json} verbatim and the guideline explicitly forbids fabricating or extrapolating results. Distilled from Peyton Jones's research-writing advice~\cite{peytonjones_writing} and the formatting requirements of NeurIPS, ICML, and ICLR.

\begin{lstlisting}[style=guideline]
# Paper Writing Guidelines

Distilled from Simon Peyton Jones, NeurIPS/ICML/ICLR formatting requirements,
and technical writing best practices.

## Core Principle

Your paper tells a story: problem → why it matters → your approach → evidence
it works → what it means. Every section serves this narrative.

## Start from proposal.md

You already wrote a research proposal (proposal.md) with introduction,
approach, related work, and references. **Use it as your foundation**:

- **Introduction**: Adapt from proposal.md's Introduction section. Add
  concrete results now that experiments are done.
- **Related Work**: Expand from proposal.md's Related Work section. Use
  the BibTeX entries in references/ for your bibliography.
- **Method**: Expand from proposal.md's Proposed Approach section. Add
  full technical details, notation, and algorithm descriptions.
- **References**: Start from the citations in proposal.md and references/.
  Add any new papers discovered during experiments.

Do NOT rewrite from scratch — refine and expand what you already have.

## Structure

Write in this order (not the order they appear in the paper):

1. Methods → Experiments → Contributions list → Conclusion
2. Then Introduction (now you know what to introduce)
3. Then Related Work
4. Abstract LAST (summarize the completed paper)

Final paper order:

```
1. Title
2. Abstract (150-250 words, one paragraph)
3. Introduction (problem, gap, contributions list, paper roadmap)
4. Related Work (funnel: broad → narrow, end with your positioning)
5. Method (complete, reproducible description)
6. Experiments (setup, results tables, ablations, analysis)
7. Discussion / Limitations
8. Conclusion
9. References
```

## Abstract

- ONE paragraph, 150-250 words
- Structure: context → problem → method → key result → implication
- Must be self-contained — readable without the rest of the paper
- No citations in the abstract
- Include one concrete quantitative result if possible

## Introduction

- Start with what is known (context)
- Identify the gap (what's missing or broken)
- State your approach (one sentence)
- List contributions explicitly:
  ```latex
  Our contributions are:
  \begin{itemize}
    \item We propose X, which addresses Y.
    \item We show that Z through experiments on A and B.
    \item We release our code and data at [URL].
  \end{itemize}
  ```
- End with a roadmap: "Section 2 reviews..., Section 3 describes..., Section 4 presents..."

## Related Work

- Organize by approach/concept, NOT chronologically
- Funnel structure: broad field → specific subproblem → directly competing methods
- For each group of related papers, explain:
  1. What they do
  2. How your work differs
- End with: "Unlike [prior work], our approach..."
- Every cited paper must be REAL and verifiable. Search Semantic Scholar
  (semanticscholar.org) to confirm papers exist before citing them.

## Method

- Complete enough that an expert can reimplement from the paper alone
- State all assumptions explicitly
- Include: model architecture, loss function, training algorithm
- Use clear notation, define every symbol on first use
- Include a method overview figure if the approach has multiple components

## Experiments

- Structure: Setup → Main results → Ablations → Analysis

### Setup subsection
- Datasets: name, size, splits, preprocessing
- Baselines: what they are, why chosen, how trained (fair comparison)
- Metrics: which ones, why appropriate
- Implementation: optimizer, lr, epochs, batch size, hardware, training time
- Seeds: how many, which values

### Results subsection
- Main comparison table with your method vs all baselines
- Bold the best value in each column
- Include ↑ or ↓ to indicate if higher/lower is better
- Every number in the paper must match experiment results in exp/ exactly

### Ablation subsection
- One table showing: full method, then remove each component
- Proves every component contributes

### Analysis subsection (optional but strengthens paper)
- Failure cases: where does your method fail and why?
- Qualitative examples: show what the model actually produces
- Training curves: show convergence behavior

## Tables

```latex
\begin{table}[t]
\caption{Comparison on [Dataset]. Best results in \textbf{bold}. ↑ means higher is better.}
\label{tab:main}
\centering
\begin{tabular}{lccc}
\toprule
Method & Accuracy ↑ & F1 ↑ & Latency (ms) ↓ \\
\midrule
Baseline A & 82.1 ± 0.3 & 79.4 ± 0.5 & 12.3 \\
Baseline B & 84.7 ± 0.2 & 81.2 ± 0.4 & 15.7 \\
\textbf{Ours} & \textbf{87.3 ± 0.2} & \textbf{84.1 ± 0.3} & 14.1 \\
\bottomrule
\end{tabular}
\end{table}
```

Rules:
- Use booktabs (\toprule, \midrule, \bottomrule) — no vertical lines
- Caption goes ABOVE the table
- Self-contained caption: readable without main text
- Reference every table in the text: "As shown in Table~\ref{tab:main}..."
- Numbers must match experiment results exactly

## Figures

```latex
\begin{figure}[t]
\centering
\includegraphics[width=0.8\linewidth]{figures/training_curve.pdf}
\caption{Training loss over epochs. Our method (blue) converges faster
than Baseline A (orange) and Baseline B (green).}
\label{fig:training}
\end{figure}
```

Rules:
- Caption goes BELOW the figure
- Self-contained caption
- Use PDF or vector format when possible (not low-res PNG)
- Readable at print size (font ≥ 8pt in the figure)
- Reference every figure: "Figure~\ref{fig:training} shows..."
- Use consistent colors across all figures

## Discussion / Limitations

- Discuss what the results mean, not just what they are
- Honestly acknowledge limitations:
  - "Our method assumes X, which may not hold in Y scenarios"
  - "We evaluated on Z datasets; generalization to other domains is untested"
- Reviewers reward honesty — hiding limitations gets papers rejected

## Conclusion

- Restate the problem and your approach (one sentence each)
- Summarize key findings with concrete numbers
- State broader implications
- Suggest future work
- DO NOT introduce new results or claims here
- 0.5-1 page

## References (CRITICAL)

- EVERY reference must be a REAL, VERIFIABLE publication
- Search Semantic Scholar (semanticscholar.org) to find and verify papers
- Fake or hallucinated citations undermine scientific integrity
- Use correct format: authors, title, venue, year
- Prefer published conference/journal papers over arXiv preprints
- Include 15-30 references for a typical ML paper
- Use \citep{} for parenthetical: "(Smith et al., 2023)"
- Use \citet{} for textual: "Smith et al. (2023) showed..."

## LaTeX Best Practices

- Use the venue's official style file (neurips_2025.sty, etc.)
- Use booktabs for tables (no vertical lines)
- Use \usepackage{hyperref} for clickable references
- Define notation with \newcommand for consistency
- Use ~ for non-breaking spaces before references: Table~\ref{tab:main}
- Compile at least twice to resolve references
- 8-10 pages for main content (excluding references and appendix)

## Common Mistakes That Get Papers Rejected

- No explicit contributions list in the introduction
- Claims not supported by evidence in the experiments
- Numbers in text don't match experiment results
- No ablation study
- Unfair baseline comparisons
- Fabricated references
- No limitations discussion
- Poor writing quality / unclear main contribution
\end{lstlisting}

\paragraph{Reviewer guideline (stage 4).}\label{app:reviewer_guideline}
Used by all three  reviewer agents (\S\ref{subsec:pr}). The ML guideline shown below is one of six per-domain reviewer guidelines; the others share the same nine 0--10 dimensions and ICLR scale, so reviewer scores are cross-domain comparable.

\begin{lstlisting}[style=guideline]
# Reviewer Guidelines

Distilled from official reviewer instructions of NeurIPS, ICML, ICLR, ACL, and TMLR.

## Your Role

You are reviewing a research paper. Your primary job is to evaluate the
scientific contribution — the novelty, soundness, significance, and clarity
of the work. Be rigorous but fair. Be specific, not vague.

You also have access to the experiment workspace (code, logs, results) for
a sanity check on results integrity.

## Overall Score (ICLR scale: 0-10, even numbers only)

| Score | Meaning |
|---|---|
| 10 | Top 5% of accepted papers, seminal paper |
| 8 | Clear accept, strong contribution |
| 6 | Marginal, needs revision |
| 4 | Below threshold, reject |
| 2 | Strong rejection, significant flaws |
| 0 | Trivial, wrong, or fabricated |

Use ONLY these values: 0, 2, 4, 6, 8, 10.
Acceptance threshold is 8. Score 6 triggers a revision loop. Score < 6 is rejected.

## Per-Dimension Scores (each 1-10)

### 1. Novelty (most important)
- Does the paper present genuinely new ideas, methods, or insights?
- **You MUST perform at least 5 distinct online searches** before assessing
  novelty. Do NOT accept the authors' novelty claims at face value.
  Required search strategies (do ALL of them):
  a) Search the exact paper title on Google Scholar and Semantic Scholar
  b) Search the core technique name + the domain (e.g., "adaptive margin metric learning")
  c) Search for each key baseline/related work cited to find papers THEY cite
  d) Search for the method's key components combined (e.g., "CLIP text encoder margin loss")
  e) Search recent proceedings (last 3 years) of the target venue for similar ideas
- If you find a paper that proposes a substantially similar method, score novelty ≤ 4
- Novel combinations of existing techniques count IF clearly reasoned
  and the combination itself provides new insight
- Incremental improvements need strong justification for why the
  increment matters
- Lack of state-of-the-art results alone does NOT justify rejection

### 2. Soundness
- Are claims well-supported by theory or experiments?
- Is the methodology appropriate for the problem?
- Are proofs correct? Is experimental design valid?
- Are assumptions stated and reasonable?
- Do the results actually support the claims made?

### 3. Significance
- Does this work matter to the community?
- Would practitioners or researchers benefit from knowing these findings?
- Does it open new research directions or solve a real problem?
- Negative results with honest analysis CAN be significant

### 4. Clarity
- Is the paper well-written and organized?
- Are contributions explicitly stated in the introduction?
- Are figures/tables self-contained with descriptive captions?
- Could an expert reproduce the work from the paper alone?
- Is the notation consistent and well-defined?

### 5. Reproducibility
- Are all hyperparameters, architectures, and training details specified?
- Is the data described (splits, sizes, preprocessing)?
- Is compute specified (hardware, runtime)?
- Are enough details provided for an independent reimplementation?

### 6. Experimental Rigor
- Are there at least 2 meaningful baselines?
- Is there an ablation study showing each component's contribution?
- Are error bars / confidence intervals reported?
- Are results from multiple runs (different seeds)?
- Are statistical significance tests used when claiming superiority?
- Are comparisons fair (same data, same compute budget)?

### 7. References
- Are key related works cited and properly discussed?
- Is the paper well-positioned relative to prior work?

### 8. Reference Integrity
- Are all references real, verifiable publications?
- **Search Semantic Scholar or Google Scholar** to verify that cited
  papers actually exist with the stated titles, authors, and venues
- Do the cited titles, authors, and venues match the actual publications?
- Are there any hallucinated or fabricated citations?

### 9. Results Integrity (sanity check — but violations mean reject)
You have access to the experiment workspace (code, logs, results.json).
You MUST verify ALL of the following:
- Read results.json and compare EVERY number in the paper's tables against it
- Check that experiment source code (.py files) exists in the workspace.
  If NO source code is present, this is a major integrity concern (score ≤ 4)
- Read experiment logs and verify they show actual training runs (epochs, losses, etc.)
- Check that the code implements what the paper describes (not a different method)
- Verify figures are generated from the actual results, not fabricated

The primary evaluation is the scientific contribution. However, any of
the following are grounds for **automatic rejection**:
- References that don't exist (fake citations)
- Experiment code that cannot run or doesn't produce the claimed results
- Logs that show different numbers than what the paper reports
- Numbers in the paper that don't match results.json
- Missing experiment source code with no explanation

These are not minor issues — they indicate the research is not trustworthy.

## Decision Guidelines

Your overall_score determines the decision:

| Score | Decision |
|---|---|
| 10 | accept |
| 8 | accept |
| 6 | revision (marginal, needs revision) |
| 4 | reject |
| 2 | reject (strong) |
| 0 | reject (fabricated/trivial) |

## Review Structure

Your review must include:
1. **Summary**: 2-3 sentence overview of what the paper does (no critique here)
2. **Novelty assessment**: What you found when searching for existing work
3. **Strengths**: Specific positives with evidence from the paper
4. **Weaknesses**: Specific issues — be constructive and actionable
5. **Detailed feedback**: How to improve the paper
6. **Questions**: Points that could change your assessment
7. **Integrity check**: Brief note on whether results appear genuine

## Common Review Mistakes to Avoid

- Don't dismiss results as "obvious in retrospect"
- Don't require SOTA results when the paper doesn't claim SOTA
- Don't reject for acknowledged limitations
- Don't demand experiments beyond the paper's stated scope
- Don't use vague criticism ("the paper is unclear") — be specific
- Don't let personal methodology preferences bias your review
- Evaluate each contribution independently, not as a bundle
- Don't conflate "I don't find this interesting" with "this is not novel"
\end{lstlisting}

\section{Per-seed paper titles and scores}
\label{app:per_seed}

Tables~\ref{tab:per_seed_claude}--\ref{tab:per_seed_kimi} list all 117 generated papers, split per agent, with the seed, trial index, full title, SAR overall score, and mean PR score (over 3 reviewers).

\begin{longtable}{p{2.6cm}cp{8.2cm}cc}
\caption{Per-seed paper titles and scores for \textbf{Claude Code}.}\label{tab:per_seed_claude}\\
\toprule
\textbf{Seed} & \textbf{Trial} & \textbf{Title} & \textbf{SAR} & \textbf{PR} \\
\midrule
\endfirsthead
\multicolumn{5}{c}{\tablename\ \thetable{} -- continued from previous page} \\
\toprule
\textbf{Seed} & \textbf{Trial} & \textbf{Title} & \textbf{SAR} & \textbf{PR} \\
\midrule
\endhead
AI for Biology & t1 & When Does Coarse-to-Fine Training Help? Ablation Insights from Curriculum Contrastive Learning for Enzyme Function Prediction & 4.50 & 4.67 \\
 & t2 & EpiGNN: Multi-Mutation Protein Fitness Prediction via Message Passing on Language Model-Derived Residue Coupling Graphs & 5.20 & 4.67 \\
 & t3 & Supervised Learning on PLM Embeddings for Multi-Mutant Protein Fitness Prediction: When Do Structural Priors Help? & 5.20 & 4.00 \\
\midrule
Causal Learning & t1 & E-Valued Causal Discovery: Constraint-Based Structure Learning with Anytime-Valid FDR Control & 5.70 & 5.33 \\
 & t2 & Know Your Assumptions: Assumption-Adaptive Edge Orientation for Robust Causal Discovery via Data-Driven Diagnostics & 5.60 & 3.33 \\
 & t3 & When Do Causal Discovery Algorithms Disagree? Diagnosing Assumption Violations via Per-Edge Profiling & 6.30 & 4.67 \\
\midrule
Compiler Optimization & t1 & The Algebra of Compiler Passes: An Empirical Study of Idempotency, Commutativity, and Convergence in LLVM Optimization Pipelines & 5.60 & 5.33 \\
 & t2 & ShapleyPass: Game-Theoretic Attribution and Interaction Analysis of Compiler Optimization Passes & 5.60 & 5.33 \\
 & t3 & ShapleyPass: Quantifying Higher-Order Interactions Among Compiler Optimization Passes via Shapley Interaction Indices & 4.50 & 4.00 \\
\midrule
Computer Vision & t1 & Entropy-Guided Adaptive Token Merging for Robust and Efficient Vision Transformers & 5.80 & 6.00 \\
 & t2 & Attention Entropy Profiling for Training-Free Out-of-Distribution Detection in Vision Transformers & 4.40 & 4.00 \\
 & t3 & Spectral Token Gating for Vision Transformer Robustness: A Negative Result with Insights on Frequency-Domain Corruption Detection in Embedding Space & 5.80 & 5.33 \\
\midrule
Data Integration \& Cleaning & t1 & Structural Sparsity in Constraint Interactions: An Empirical Study of Multi-Constraint Data Repair Decomposition & 5.20 & 3.33 \\
 & t2 & Error Amplification in Entity Resolution Pipelines: A Formal Analysis of Stage-Wise Quality Propagation & 3.90 & 5.33 \\
 & t3 & Characterizing Operator Interaction Effects in Data Cleaning Pipelines & 5.60 & 5.33 \\
\midrule
Datasets \& Benchmarks & t1 & FlipBench: Measuring Directional Reasoning Asymmetry in Large Language Models & 5.60 & 4.67 \\
 & t2 & consistbench\{ & 6.30 & 5.33 \\
 & t3 & SkillStack: A Procedurally Generated Benchmark for Measuring Compositional Cognitive Skill Gaps in Large Language Models & 5.80 & 4.67 \\
\midrule
Generative Models & t1 & Spectral Consistency Distillation: Frequency-Adaptive Teacher Supervision for Few-Step Flow Matching & 3.10 & 4.67 \\
 & t2 & Prediction Coherence is Not a Quality Signal: A Negative Result on Verifier-Free Inference-Time Scaling for Diffusion Models & 6.30 & 5.33 \\
 & t3 & Conditioning-Space Guidance for Diffusion Transformers: When Does Single-Pass Classifier-Free Guidance Work? & 6.30 & 4.67 \\
\midrule
Interpretability of Learned Repr. & t1 & The Convergent Core: Connecting Seed Stability, Cross-Model Universality, and Causal Importance of Sparse Autoencoder Features & 5.60 & 4.00 \\
 & t2 & The Functional Anatomy of Sparse Features in Language Models & 5.80 & 5.33 \\
 & t3 & Faithful by Consensus: Identifying Causally Important Features Through Multi-Seed Sparse Autoencoder Agreement & 5.80 & 4.00 \\
\midrule
Natural Language Processing & t1 & Context-Contrastive Uncertainty Decomposition for Reliable Retrieval-Augmented Generation & 6.10 & 4.00 \\
 & t2 & SpecCheck: Testing Confidence Monotonicity Across Specificity Levels for LLM Hallucination Detection---A Negative Result & 5.80 & 4.67 \\
 & t3 & Know When to Look: Parametric-Retrieval Agreement as a Calibration Signal for Retrieval-Augmented Generation & 5.80 & 5.33 \\
\midrule
Operating System Design & t1 & MarkovTier: Anticipatory Page Migration via Markov Phase Models for Tiered Memory Systems & 6.10 & 4.67 \\
 & t2 & The Bandwidth Knapsack: Optimal Migration Scheduling for Tiered Memory Systems & 5.20 & 4.67 \\
 & t3 & Invisible Cycles: Characterizing and Quantifying CPU Time Displacement from Asynchronous Kernel Execution in Modern Linux & 6.10 & 4.00 \\
\midrule
Privacy in ML & t1 & MemPrune: Investigating Gradient Dispersion as a Memorization-Aware Neural Network Pruning Criterion & 4.90 & 3.33 \\
 & t2 & Difficulty-Calibrated Unlearning Auditing: Exposing Per-Sample Privacy Gaps in Machine Unlearning & 5.20 & 3.33 \\
 & t3 & The Compounding Cost: How Differential Privacy and Model Compression Jointly Amplify Fairness Degradation & 5.60 & 4.67 \\
\midrule
Probabilistic Methods & t1 & Confidence Sequences for Markov Chain Monte Carlo: Anytime-Valid Estimation with Sequential Guarantees & 5.20 & 5.33 \\
 & t2 & Optimal Error Budgeting for Heterogeneous Sketch Pipelines in Approximate Stream Processing & 6.10 & 5.33 \\
 & t3 & Sublinear-Memory Confidence Sequences for Streaming Quantiles & 5.70 & 5.33 \\
\midrule
Supervised Repr. Learning & t1 & Confusion-Geometric Supervised Contrastive Learning: Shaping Embedding Geometry from Training Dynamics & 4.40 & 4.00 \\
 & t2 & The Neural Collapse--Calibration Connection is Dataset-Dependent: An Empirical Investigation via Controlled Within-Class Geometry & 5.80 & 4.00 \\
 & t3 & Confusion-Calibrated Supervised Contrastive Learning: Adaptive Class-Pair Reweighting from Training Dynamics & 5.00 & 3.33 \\
\bottomrule
\end{longtable}

\begin{longtable}{p{2.6cm}cp{8.2cm}cc}
\caption{Per-seed paper titles and scores for \textbf{Codex}.}\label{tab:per_seed_codex}\\
\toprule
\textbf{Seed} & \textbf{Trial} & \textbf{Title} & \textbf{SAR} & \textbf{PR} \\
\midrule
\endfirsthead
\multicolumn{5}{c}{\tablename\ \thetable{} -- continued from previous page} \\
\toprule
\textbf{Seed} & \textbf{Trial} & \textbf{Title} & \textbf{SAR} & \textbf{PR} \\
\midrule
\endhead
AI for Biology & t1 & Does Systema-Style Perturbed-Reference Residualization Help as a Training Target for Unseen Single-Cell Perturbation Prediction? A Pre-Registered Benc & 5.00 & 4.00 \\
 & t2 & Masked-Child Surrogate Calibration for Safe EC Prefix Decisions: A Leakage-Controlled Benchmark for Future-Child Emergence & 5.00 & 4.67 \\
 & t3 & SPARE-Gain: A Low-Compute Benchmark of Baseline-Relative Routing for Unseen-Perturbation Pseudobulk Prediction & 4.50 & 4.67 \\
\midrule
Causal Learning & t1 & Benchmarking Subset Aggregation for Classical Causal Discovery Under Marginalization Error & 5.00 & 4.67 \\
 & t2 & PACER-Cert as a Benchmark for Stopping-Certificate Calibration in CPU-Only Active Causal Discovery & 5.00 & 4.67 \\
 & t3 & When Do Path-Dependent Setup Costs Matter in Sequential Causal Design? & 3.20 & 4.67 \\
\midrule
Compiler Optimization & t1 & DebtAware Beyond LastRunTracking? A Proxy Feasibility Boundary Study of Typed Rerun Suppression in LLVM & 2.70 & 4.67 \\
 & t2 & Typed Skip-Versus-Global Scheduling for LLVM Cleanup Pipelines: A Negative Proxy Study & 5.00 & 4.00 \\
 & t3 & Optimization Remarks as a Feasibility Signal for Low-Budget LLVM Micro-Search A Pilot Against Random and ProbeDelta & 3.60 & 4.00 \\
\midrule
Computer Vision & t1 & Do Corruption-Family Text Residuals Help Zero-Shot CLIP? A Controlled Negative Result on CIFAR-C & 5.00 & 4.00 \\
 & t2 & FOCUS: Object-Centric Evidence as a Causal Update Gate for Realistic Online Vision-Language Adaptation & 5.80 & 4.67 \\
 & t3 & Object Units or Pixels? A Negative Proxy Feasibility Study for Online Segmentation Adaptation & 3.90 & 4.67 \\
\midrule
Data Integration \& Cleaning & t1 & A Preliminary Artifact-Backed Cautionary Study of Benchmark-Conditional Admissibility for Robustness Evaluation in Schema and Entity Matching & 3.90 & 4.67 \\
 & t2 & CanopyER: A Short Systems Note on Budgeted Rewrite-vs-Match Scheduling for Progressive Entity Matching & 3.80 & 4.67 \\
 & t3 & StressAudit-SM: A Compact Robustness Audit for Schema Matching Under Metadata Stress & 5.00 & 4.00 \\
\midrule
Datasets \& Benchmarks & t1 & DriftAnswer-Py: A 12-Item Executable Pilot of Accepted Python Stack Overflow Answers Under Documented Library Drift & 6.30 & 4.67 \\
 & t2 & RevisionBench: A Reproducible Failed-Construction Case Study for Abstract-Local Scientific Claim Updates & 6.20 & 4.00 \\
 & t3 & TwinBench: A Synthetic Procedural-Core Pilot for Coupled Invariance and Boundary Sensitivity in QA & 6.30 & 4.67 \\
\midrule
Generative Models & t1 & When Assignment Matters: A Pilot Study of DAAM-Assisted Compositional Text-to-Image Reranking & 4.40 & 4.00 \\
 & t2 & Decomposed Early Ranking Targets Under a Local Surrogate Evaluator for Compositional Text-to-Image Generation & 4.50 & 4.00 \\
 & t3 & bf ParaDG: An Exploratory Negative Study of Disagreement-Gated Paraphrase Blending for Text-to-Image Diffusion & 5.80 & 4.67 \\
\midrule
Interpretability of Learned Repr. & t1 & Do Shared Decoders Improve Prototype-Edit Reusability on Frozen CLIP Features? & 5.00 & 4.00 \\
 & t2 & Pair-Supervised Regularization for Selective Counterfactual Edits in Frozen Vision SAEs & 5.00 & 4.67 \\
 & t3 & Benchmarking Weakly Supervised Factor-Localized Sparse Autoencoders on Frozen Vision Features & 5.80 & 4.67 \\
\midrule
Natural Language Processing & t1 & When Does Clarification Supervision Transfer to Formal Reasoning? A Controlled Pilot of Validator-Clean vs. Matched Noisy Missing-Fact Tuning & 4.90 & 4.00 \\
 & t2 & LIMS-RAG: Localized Minimal-Support Perturbations Are a Weak but Measurable Feature Family for Sentence-Level RAG Verification & 5.00 & 4.67 \\
 & t3 & LateBind: Controlled Additive-Value Tests for Timing-Aware Shortcut Mitigation in Text Classification & 6.30 & 4.67 \\
\midrule
Operating System Design & t1 & Replay-Scoped Evidence for Bias-Corrected Counterfactual Policy Ranking in One Shared Linux Page Cache & 5.80 & 4.67 \\
 & t2 & ShareArb: Evictor-Side Responsibility for Shared Linux Page-Cache Arbitration & 5.80 & 4.67 \\
 & t3 & ShadowCache: A Simulator-Backed Trace Study of How Much Observable State Policy Ranking Needs & 5.00 & 4.67 \\
\midrule
Privacy in ML & t1 & Who Was in the Recent Window? A Rigorous Audit of Online Test-Time Adaptation Privacy & 5.20 & 5.33 \\
 & t2 & Matched-Budget Evaluation of Weak-View Residuals for One-Run Differential Privacy Auditing & 4.70 & 4.00 \\
 & t3 & Are Early Artifact Forecasts Actionable for Membership Privacy? A Budget-Matched Study of Selective Intervention & 3.90 & 4.67 \\
\midrule
Probabilistic Methods & t1 & CoSBC: Dependence-Specialized Enriched SBC with Symmetric Pooled Ranking & 4.50 & 4.67 \\
 & t2 & Hierarchical Diagonal-GMM Posteriors for Localized Conformal Prediction: A Scoped Negative Result & 5.00 & 4.67 \\
 & t3 & How Far Does Probe-Only Recalibration Transfer in Bayesian Quadrature? & 5.00 & 4.67 \\
\midrule
Supervised Repr. Learning & t1 & STRIDE: Reliability-Gated Class-Relation Smoothing for Self-Supervised Transfer & 4.50 & 4.67 \\
 & t2 & Adaptive Prototype Granularity in Frozen-Feature Contrastive Adaptation & 5.00 & 4.67 \\
 & t3 & How Much Signal Is in Early Training Trajectories? A Matched-Budget Study of Pseudo-Group Inference & 5.80 & 5.33 \\
\bottomrule
\end{longtable}

\begin{longtable}{p{2.6cm}cp{8.2cm}cc}
\caption{Per-seed paper titles and scores for \textbf{Kimi Code}.}\label{tab:per_seed_kimi}\\
\toprule
\textbf{Seed} & \textbf{Trial} & \textbf{Title} & \textbf{SAR} & \textbf{PR} \\
\midrule
\endfirsthead
\multicolumn{5}{c}{\tablename\ \thetable{} -- continued from previous page} \\
\toprule
\textbf{Seed} & \textbf{Trial} & \textbf{Title} & \textbf{SAR} & \textbf{PR} \\
\midrule
\endhead
AI for Biology & t1 & contextstab: Context-Aware Protein Stability Prediction Using Real Single-Cell Transcriptomics & 4.40 & 4.00 \\
 & t2 & Tri-Con: Tri-Hierarchy Contrastive Learning for Cell Ontology-Guided Single-Cell Analysis & 4.40 & 3.33 \\
 & t3 & CellStratCP: Cell-Type-Stratified Adaptive Conformal Prediction for Calibrated Uncertainty in Single-Cell RNA-seq Imputation & 5.20 & 5.33 \\
\midrule
Causal Learning & t1 & AIT-LCD: Adaptive Information-Theoretic Local Causal Discovery with Explicit Conditioning Set Awareness & 4.10 & 2.67 \\
 & t2 & SPICED: Structural Prior Integration for Constrained Estimation of Directed Information & 5.20 & 4.00 \\
 & t3 & On the Challenges of Multi-Fidelity Conditional Independence Testing for Causal Discovery: An Empirical Study & 2.00 & 4.00 \\
\midrule
Compiler Optimization & t1 & From Branches to Bytes: A Negative Result in Extending Learned Static Prediction to Data Layout Optimization & 4.40 & 4.00 \\
 & t2 & LEOPARD: Lightweight Learned Guidance for Equality Saturation in Compiler Optimization & 4.40 & 2.00 \\
 & t3 & Joint Compute and Layout Optimization via Hierarchical E-Graphs & 4.40 & 2.67 \\
\midrule
Computer Vision & t1 & DU-VPT: Decomposed Uncertainty-Guided Visual Prompt Tuning for Test-Time Adaptation & 4.40 & 2.67 \\
 & t2 & Adaptive Prototype-Aware Consistency with Learnable Augmentation Policies for Single-Image Test-Time Adaptation & 3.60 & 3.33 \\
 & t3 & CASS-ViM: Content-Adaptive Selective Scanning for Vision State Space Models & 4.40 & 4.00 \\
\midrule
Data Integration \& Cleaning & t1 & CESF: A Controllable Error Synthesis Framework for Reproducible Data Cleaning Evaluation & 3.90 & 3.33 \\
 & t2 & Towards LLM-as-Compiler for Data Cleaning: A Feasibility Study & 4.10 & 4.00 \\
 & t3 & CleanBP: Making Belief Propagation Practical for Holistic Data Repair via FD-Specific Sparsification & 4.10 & 4.00 \\
\midrule
Datasets \& Benchmarks & t1 & textbf\{CompViz: A Dynamic Benchmark for Compositional Visual Reasoning with Sub-100ms Generation & 5.20 & 3.33 \\
 & t2 & IntrospectBench: A Cross-Domain Benchmark for Evaluating Step-Level Reasoning Introspection in Large Language Models & 5.20 & 2.67 \\
 & t3 & DynaScale: Dynamic Difficulty Scaling for Maintaining Discriminative Power in AI Benchmarks & 5.20 & 4.00 \\
\midrule
Generative Models & t1 & Flow-Guided Token Routing: Adaptive Computation Allocation for Efficient Flow Matching & 3.70 & 2.00 \\
 & t2 & Distance-Aware Flow Matching for LiDAR Point Cloud Generation & 4.90 & 3.33 \\
 & t3 & VAST: Velocity-Adaptive Spatially-varying Timesteps for Training-Free Acceleration of Diffusion Models & 4.10 & 2.00 \\
\midrule
Interpretability of Learned Repr. & t1 & CAGER: Causal Geometric Explanation Recovery A Framework for Grounding Interpretability in Causal Subspace Geometry & 4.40 & 4.00 \\
 & t2 & Intervention Fidelity Scoring: Characterizing the Precision-Magnitude Trade-off in SAE-Based Steering & 3.70 & 3.33 \\
 & t3 & PhaseMine: Detecting Feature Emergence Phase Transitions via Dynamic Sparse Probing & 4.40 & 4.00 \\
\midrule
Natural Language Processing & t1 & SAE-GUIDE: Sparse Autoencoder-Guided Uncertainty-aware Information Detection and Enhancement for Multi-Hop Retrieval & 4.10 & 2.67 \\
 & t2 & Confidence-Dynamic Heterogeneous Reasoning: protect A Study of Challenges in Adaptive Strategy Selection & 3.30 & 2.67 \\
 & t3 & Entropy-Guided Stepwise Revision: In-Chain Self-Correction for Efficient Reasoning & 3.60 & 3.33 \\
\midrule
Operating System Design & t1 & UniSched: A Critical Analysis of Simulation-Based Evaluation for CXL-Aware CPU Scheduling & 4.40 & 3.33 \\
 & t2 & WattSched: Adaptive Workload-Aware Energy Scheduling for Heterogeneous Multi-Core Systems using sched\_ext & 4.40 & 4.00 \\
 & t3 & KAPHE: Kernel-Aware Performance Heuristic Extraction vspace\{0.2em & 3.60 & 4.67 \\
\midrule
Privacy in ML & t1 & Post-Hoc Compression-Aware Differential Privacy: Optimizing DP Training for Deployed Compressed Models & 5.30 & 0.00 \\
 & t2 & textbf\{G3P: Gradient-Guided Privacy-Preserving Pruning via Train-Test Gradient Saliency & 4.40 & 4.00 \\
 & t3 & On the Limitations of Gradient-Based Verification for Machine Unlearning & 5.20 & 4.00 \\
\midrule
Probabilistic Methods & t1 & Decaying HyperLogLog: Continuous-Time Cardinality Estimation with Exponential Aging & 2.40 & 4.00 \\
 & t2 & Streaming Multi-Scale Adaptive Kernel Conformal Prediction & 5.20 & 4.00 \\
 & t3 & Comparative Analysis of Adaptation Criteria for Gradient-Based Discrete MCMC: When Acceptance-Rate Trumps Jump-Distance & 5.20 & 4.67 \\
\midrule
Supervised Repr. Learning & t1 & Feature-Diversity-Aware Supervised Contrastive Learning: Mitigating Feature Suppression through Adaptive Pair Weighting & 5.20 & 2.00 \\
 & t2 & Gradient-Confusion Aware Supervised Contrastive Learning & 2.80 & 4.00 \\
 & t3 & ETF-SCL: Equiangular Tight Frame Guided Supervised Contrastive Learning for Long-Tail Recognition & 2.50 & 2.67 \\
\bottomrule
\end{longtable}

\section{Per-domain SAR and PR breakdown}
\label{app:per_domain}

Table~\ref{tab:per_domain} reports SAR and PR mean scores for each of the 13 research domains, averaged across the three agents ($n=9$ papers per domain). Domains are ordered by SAR within each platform.

\begin{table}[h]
\centering
\caption{Per-domain SAR and PR mean scores, averaged across the three agents.}
\label{tab:per_domain}
\small
\begin{tabular}{llccc}
\toprule
\textbf{Platform} & \textbf{Domain}                  & \textbf{SAR} & \textbf{PR} & \textbf{SAR$-$PR} \\
\midrule
\multirow{5}{*}{CPU} & Operating System Design        & 5.16 & 4.37 & 0.79 \\
                     & Probabilistic Methods          & 4.92 & 4.74 & 0.18 \\
                     & Causal Learning                & 4.68 & 4.22 & 0.46 \\
                     & Compiler Optimization          & 4.47 & 4.00 & 0.47 \\
                     & Data Integration \& Cleaning   & 4.39 & 4.30 & 0.09 \\
\midrule
\multirow{8}{*}{GPU} & Datasets \& Benchmarks         & 5.79 & 4.22 & 1.57 \\
                     & Interpretability of Reps       & 5.06 & 4.22 & 0.84 \\
                     & NLP                            & 4.99 & 4.00 & 0.99 \\
                     & Privacy in ML                  & 4.93 & 3.70 & 1.23 \\
                     & AI for Biology                 & 4.82 & 4.37 & 0.45 \\
                     & Computer Vision                & 4.79 & 4.30 & 0.49 \\
                     & Generative Models              & 4.79 & 3.85 & 0.94 \\
                     & Supervised Repr.\ Learning     & 4.56 & 3.85 & 0.71 \\
\bottomrule
\end{tabular}
\end{table}

Figure~\ref{fig:per_domain_breakdown} (in \S\ref{subsec:cpu_gpu_gap}) visualises the per-domain PR scores; the parallel SAR view is Figure~\ref{fig:per_domain_sar} below. Table~\ref{tab:cpu_gpu} reports the same data collapsed to the CPU/GPU split per agent. Figure~\ref{fig:score_heatmap} gives the per-(seed, trial) score grids for both SAR and PR alongside each agent.

\begin{table}[h]
\centering
\caption{SAR vs.\ PR by agent and compute platform.}
\label{tab:cpu_gpu}
\small
\begin{tabular}{lccccc}
\toprule
\textbf{Agent}        & \textbf{Platform} & \textbf{n} & \textbf{SAR} & \textbf{PR} & \textbf{SAR$-$PR} \\
\midrule
\multirow{2}{*}{Claude Code} & CPU & 15 & 5.49 & 4.76 & 0.74 \\
                             & GPU & 24 & 5.42 & 4.50 & 0.92 \\
\midrule
\multirow{2}{*}{Codex}       & CPU & 15 & 4.55 & 4.53 & 0.02 \\
                             & GPU & 24 & 5.16 & 4.50 & 0.66 \\
\midrule
\multirow{2}{*}{Kimi Code}   & CPU & 15 & 4.12 & 3.69 & 0.43 \\
                             & GPU & 24 & 4.32 & 3.19 & 1.12 \\
\bottomrule
\end{tabular}
\end{table}

\begin{figure}[h]
  \centering
  \includegraphics[width=\linewidth]{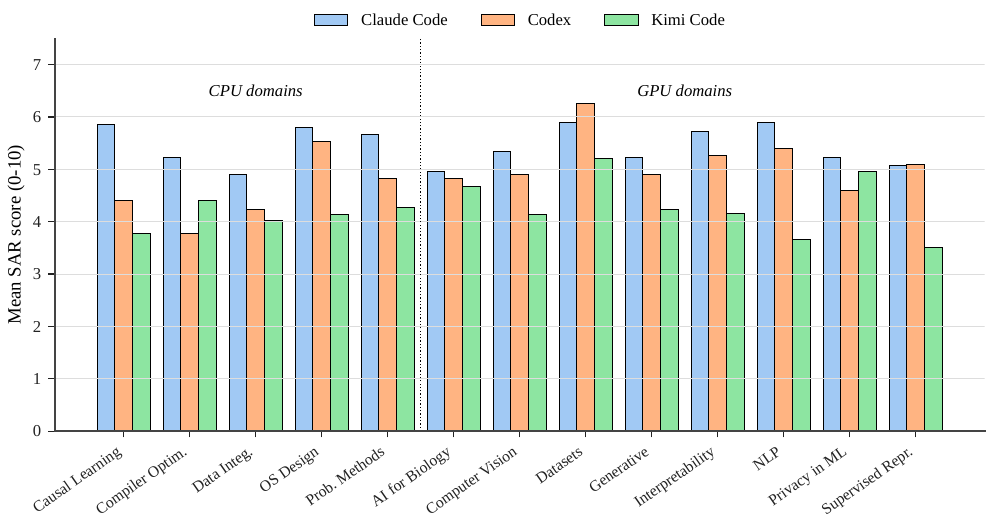}
  \caption{Per-domain mean SAR scores by agent (companion to Figure~\ref{fig:per_domain_breakdown}).}
  \label{fig:per_domain_sar}
\end{figure}

\begin{figure}[h]
  \centering
  \includegraphics[width=\linewidth]{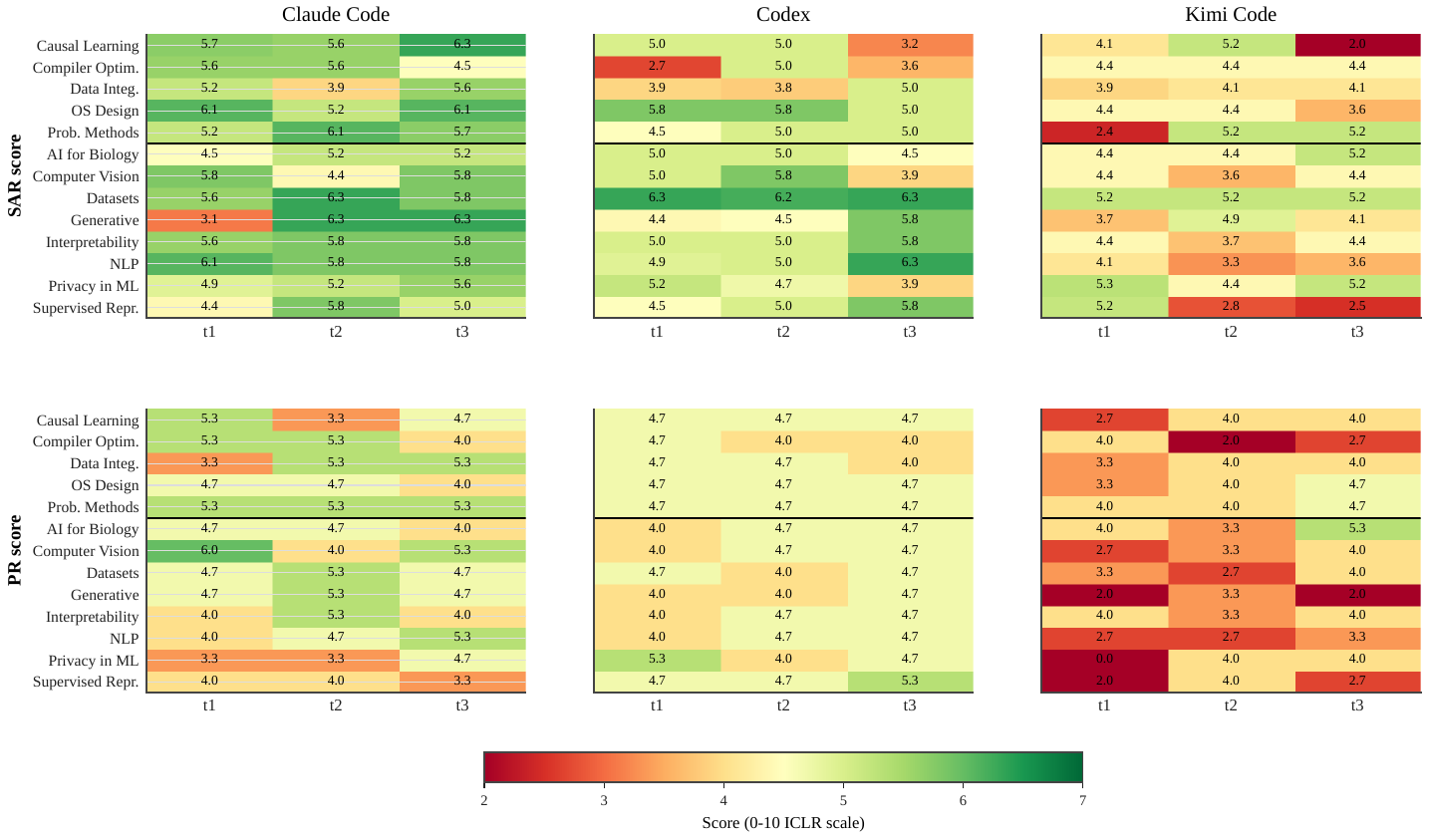}
  \caption{Per-(seed, trial) SAR (top) and PR (bottom) score heatmaps per agent. CPU seeds above the divider, GPU below; columns are trial indices ($t_1$, $t_2$, $t_3$).}
  \label{fig:score_heatmap}
\end{figure}



\section{Time analysis}
\label{app:time}

\begin{figure}[h]
  \centering
  \includegraphics[width=\linewidth]{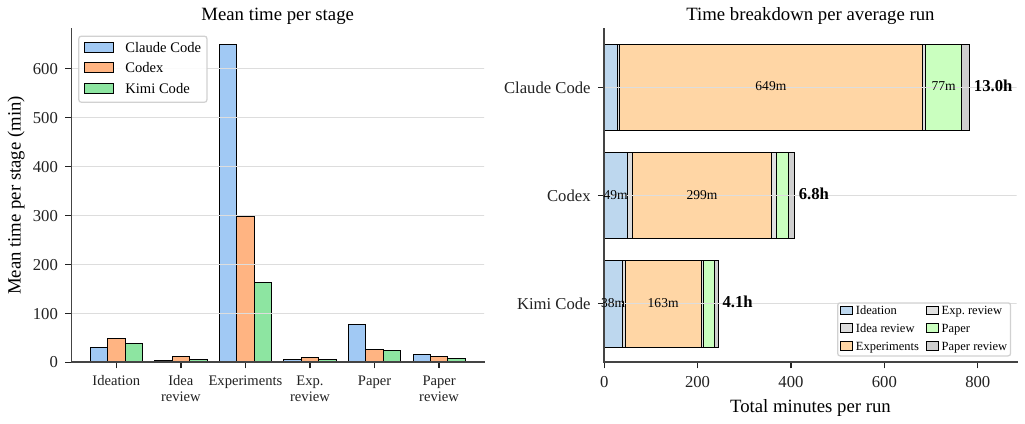}
  \caption{Mean wall-clock per pipeline stage (left, in minutes) and total time per average run with stages stacked (right, in minutes), for each agent.}
  \label{fig:time_per_stage_full}
\end{figure}

\paragraph{Per-stage time.} Figure~\ref{fig:time_per_stage_full} reports per-agent wall-clock time per pipeline stage in two views: grouped means per stage (left) and a stacked breakdown of total minutes per average run (right). Experiments dominate (67--83\% of total) and self-refinement is only 3--8\%. The same totals (Claude Code 13.0\,h, Codex 6.8\,h, Kimi Code 4.1\,h) appear as the right panel of Figure~\ref{fig:pipeline_stats} in \S\ref{subsec:pipeline_stats}.

\begin{figure}[h]
  \centering
  \includegraphics[width=\linewidth]{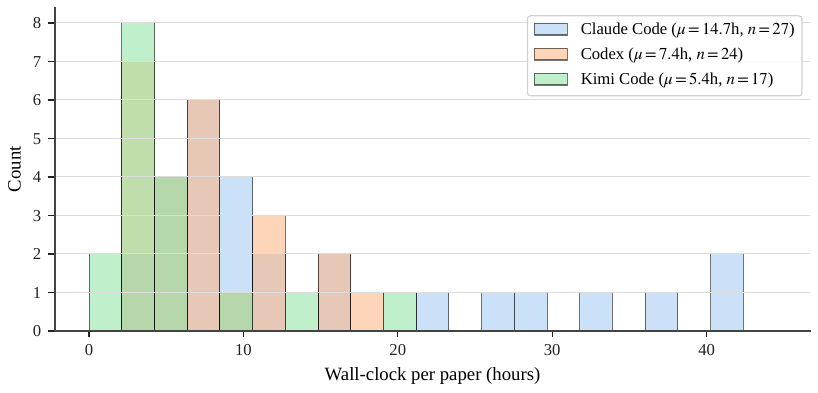}
  \caption{Per-paper total wall-clock distribution per agent, from \texttt{tracker.json} logs.}
  \label{fig:wall_time}
\end{figure}

\paragraph{Per-paper distribution.} Figure~\ref{fig:wall_time} shows the per-paper total wall-clock distribution from the released \texttt{tracker.json} logs. Claude Code's distribution has the longest tail ($>$\,40\,h on some runs, driven by experiment-execution self-refinement loops); Codex and Kimi Code are tighter and shorter.

\section{H100 scaling experiment (per-domain breakdown)}
\label{app:h100}

We re-ran all 8 GPU seeds with Codex on 8$\times$ NVIDIA H100 (80~GB) for 3 trials each, with budget matched to the A6000 runs (\S\ref{subsec:compute_bottleneck}). Table~\ref{tab:h100} reports per-domain Codex PR scores on H100 vs.\ A6000. The aggregate change is small ($-0.24$ on average) and agent-level SAR also drops, indicating that compute is not the binding constraint.

\begin{table}[h]
\centering
\caption{Per-domain Codex PR mean scores on 8$\times$ NVIDIA H100 vs.\ 1$\times$ A6000, budget-matched.}
\label{tab:h100}
\small
\begin{tabular}{lccc}
\toprule
\textbf{GPU domain seed}                & \textbf{Codex H100 PR} & \textbf{Codex A6000 PR} & \textbf{$\Delta$} \\
\midrule
AI for Biology                          & 4.89                   & 4.67                    & $+0.22$ \\
Interpretability of Learned Repr.\      & 4.78                   & 4.67                    & $+0.11$ \\
Computer Vision                         & 4.33                   & 4.33                    & $+0.00$ \\
Natural Language Processing             & 4.22                   & 4.33                    & $-0.11$ \\
Privacy in ML                           & 4.22                   & 4.00                    & $+0.22$ \\
Supervised Repr.\ Learning              & 4.22                   & 4.67                    & $-0.45$ \\
Generative Models                       & 3.78                   & 4.33                    & $-0.55$ \\
Datasets and Benchmarks                 & 3.67                   & 4.67                    & $-1.00$ \\
\midrule
Average                                 & 4.26                   & 4.50                    & $-0.24$ \\
\bottomrule
\end{tabular}
\end{table}

\section{Reviewer analysis}
\label{app:selfreview}

\paragraph{Self-refinement effectiveness.} At each of the three authoring stages (ideation, experiment, paper), the agent self-reviews and revises if the score falls below threshold (up to 3 rounds). Table~\ref{tab:selfreview} reports, for each (agent, stage), the share of revision rounds in which the score \emph{improved}, stayed the \emph{same}, or \emph{dened}, plus the mean delta. Self-refinement is effective for ideation and experiment-execution (avg.\ $+2.0$ to $+3.0$ for Claude Code / Kimi Code at those gates), but limited for paper writing, where revising tends to leave scores unchanged or lower them.

\begin{table}[!ht]
\centering
\caption{Self-refinement effectiveness per agent and gate (across all revision rounds).}
\label{tab:selfreview}
\small
\begin{tabular}{lcccc}
\toprule
\textbf{Agent / Gate} & \textbf{Improved} & \textbf{Same} & \textbf{Dened} & \textbf{Avg $\Delta$} \\
\midrule
Claude Code / Idea         & \textbf{100\%} &  0\% &  0\% & $+2.1$ \\
Claude Code / Experiment   &  88\%          &  8\% &  4\% & $+2.2$ \\
Claude Code / Paper        &  43\%          & 34\% & 23\% & $+0.0$ \\
\midrule
Codex / Idea               &  35\%          & 51\% & 15\% & $+0.4$ \\
Codex / Experiment         &  78\%          & 20\% &  2\% & $+2.0$ \\
Codex / Paper              &  64\%          & 29\% &  7\% & $+1.5$ \\
\midrule
Kimi Code / Idea           & \textbf{100\%} &  0\% &  0\% & $+3.0$ \\
Kimi Code / Experiment     &  91\%          &  9\% &  0\% & $+2.2$ \\
Kimi Code / Paper          &  61\%          & 34\% &  5\% & $+1.6$ \\
\bottomrule
\end{tabular}
\end{table}

\begin{figure}[!ht]
\centering
\includegraphics[width=\linewidth]{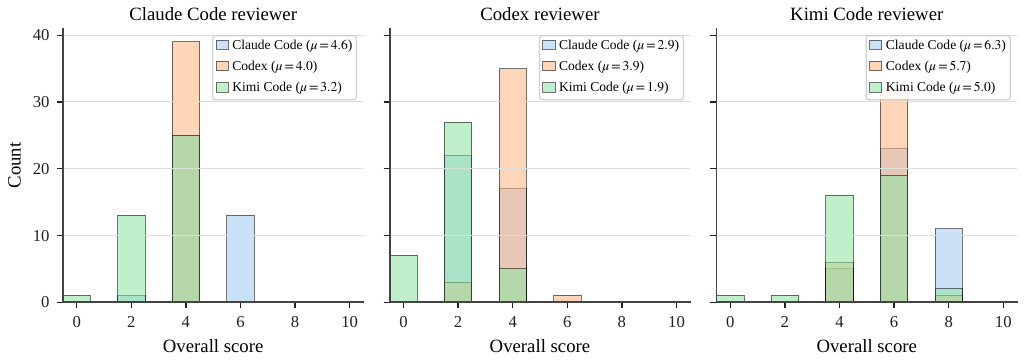}
\caption{Score distribution by (reviewer, reviewee) across all 117 papers. Codex is the strictest reviewer; Kimi Code is the most lenient. Reviewer effects dominate over self-favouring.}
\label{fig:reviewer_bias}
\end{figure}

\paragraph{Peer-review bias.} Each paper is scored by all three  agents acting as reviewers. Reviewer severity differs sharply (Table~\ref{tab:reviewer_severity}, 2.8-point mean spread between strictest and most lenient), and the reviewer-by-reviewee matrix in Table~\ref{tab:reviewer_matrix} and Figure~\ref{fig:reviewer_bias} shows that the bias is mostly an across-the-board reviewer effect rather than an agent-specific self-favouring effect: Codex gives every reviewee its lowest scores, Kimi Code gives every reviewee its highest scores. Notably, Kimi Code does \emph{not} score its own papers highest (5.0), but is still substantially more lenient overall, including on agents that produce stronger papers. Single-reviewer self-evaluation therefore drifts toward the reviewer's own bias, motivating the triple-reviewer protocol used throughout the paper.

\begin{table}[!ht]
\centering
\caption{Per-reviewer severity: distribution of overall scores each  agent assigns across all 117 papers it reviews.}
\label{tab:reviewer_severity}
\small
\begin{tabular}{lcccc}
\toprule
\textbf{Reviewer} & \textbf{n} & \textbf{mean} & \textbf{$\sigma$} & \textbf{range} \\
\midrule
Codex             & 117 & \textbf{2.89} & 1.24 & 0--6 \\
Claude Code       & 117 & 3.95          & 1.03 & 0--6 \\
Kimi Code         & 117 & \textbf{5.69} & 1.32 & 0--8 \\
\bottomrule
\end{tabular}
\end{table}

\begin{table}[!ht]
\centering
\caption{Mean PR score by (reviewer, reviewee). Rows are reviewers; columns are paper authors.}
\label{tab:reviewer_matrix}
\small
\begin{tabular}{lcccc}
\toprule
\textbf{Reviewer} $\downarrow$ / \textbf{Authored by} $\rightarrow$ & \textbf{Claude Code} & \textbf{Codex} & \textbf{Kimi Code} & \textbf{Avg given} \\
\midrule
Claude Code & 4.6           & 4.0           & 3.2           & 3.9 \\
Codex       & 2.9           & 3.9           & 1.9           & \textbf{2.9} \\
Kimi Code   & \textbf{6.2}  & \textbf{5.7}  & \textbf{5.0}  & \textbf{5.7} \\
\bottomrule
\end{tabular}
\end{table}

\section{Manually annotated SAR final-decision acceptance rates}
\label{app:manual_sar_decisions}

SAR's continuous 0--10 score does not map directly to accept/reject. We manually inspected every SAR review across the 117 agent-generated papers, the 200 ICLR 2025 papers, and the 102 FARS papers, and assigned a binary decision from the verbal recommendation (treating ``borderline accept'', ``accept with revision'', and ``conditional accept'' as accepts). The ICLR-Weighted row in Table~\ref{tab:manual_accept_rates} mixes accepted and rejected rates in proportion to ICLR's $\sim$32\% acceptance rate.

\begin{table}[h]
\centering
\caption{Manually annotated SAR final-decision acceptance rates per system.}
\label{tab:manual_accept_rates}
\small
\begin{tabular}{lcc}
\toprule
\textbf{System}                          & \textbf{n}  & \textbf{Accept \%} \\
\midrule
ICLR 2025 Accepted (human)               & 100         & 76.0\% \\
ICLR 2025 Weighted (32\% acc / 68\% rej) & 200         & 59.7\% \\
ICLR 2025 Rejected (human)               & 100         & 52.0\% \\
\midrule
\textbf{Claude Code}                     &  39         & \textbf{41.0\%} \\
FARS                                     & 102         & 21.6\% \\
Codex                                    &  39         & 12.8\% \\
Kimi Code                                &  39         & 5.1\% \\
\bottomrule
\end{tabular}
\end{table}

\section{Case studies}
\label{app:cases}

We illustrate the failure modes documented in \S\ref{sec:limitations} with six representative papers.

\subsection{Case 1: When Do Causal Discovery Algorithms Disagree? Diagnosing Assumption Violations via Per-Edge Profiling}
\label{case:1}
\textit{Claude Code can occasionally produce a paper with a real insight, but weak evidence still prevents it from being convincing.} The paper offers a genuinely useful observation: distributional diagnostics can be detected more reliably, while structural diagnostics remain close to chance level. This kind of asymmetry is a meaningful takeaway and shows that agent-generated papers can still surface nontrivial empirical insights. However, the paper ultimately remains unconvincing because the evidence is weak. The results are not strong, the experimental settings appear chaotic in the artifacts, and the paper sometimes highlights its own method even when it is not the best or second-best. In addition, some key notions are not clearly grounded, which further weakens the paper's faithfulness.
\begin{figure}[!ht]
  \centering
  \includegraphics[width=\textwidth]{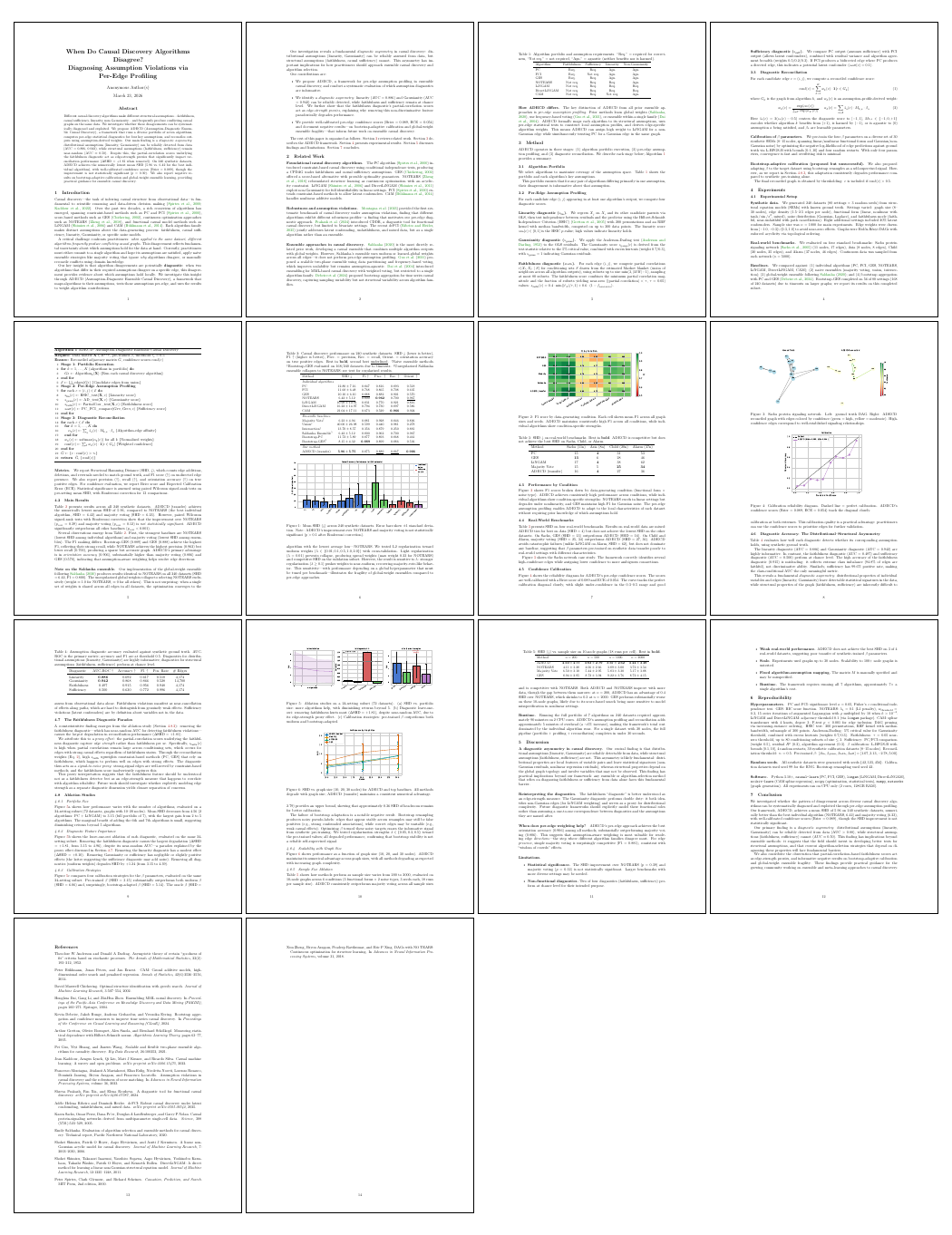}
  \caption{Page-level thumbnail of the Case 1 paper.}
  \label{fig:case1_strip}
\end{figure}

\subsection{Case 2: The Algebra of Compiler Passes: An Empirical Study of Idempotency, Commutativity, and Convergence in LLVM Optimization Pipelines}
\label{case:2}
\textit{Claude Code may overclaim or present unsupported results when experiments are weak.} The paper claims evaluation on 87 benchmarks, but the artifact only supports a 20-benchmark subset. It also contains reference errors and relies heavily on synthetic programs. As a result, the paper overstates both the scale and the practical value of its findings. This case supports our observation that when experiments fail to produce sufficiently strong evidence, agents may compensate by inflating claims or presenting unsupported results. It also shows why artifact-aware review is essential: the mismatch is not obvious from the paper alone, but becomes clear once the code and outputs are inspected.
\begin{figure}[!ht]
  \centering
  \includegraphics[width=\textwidth]{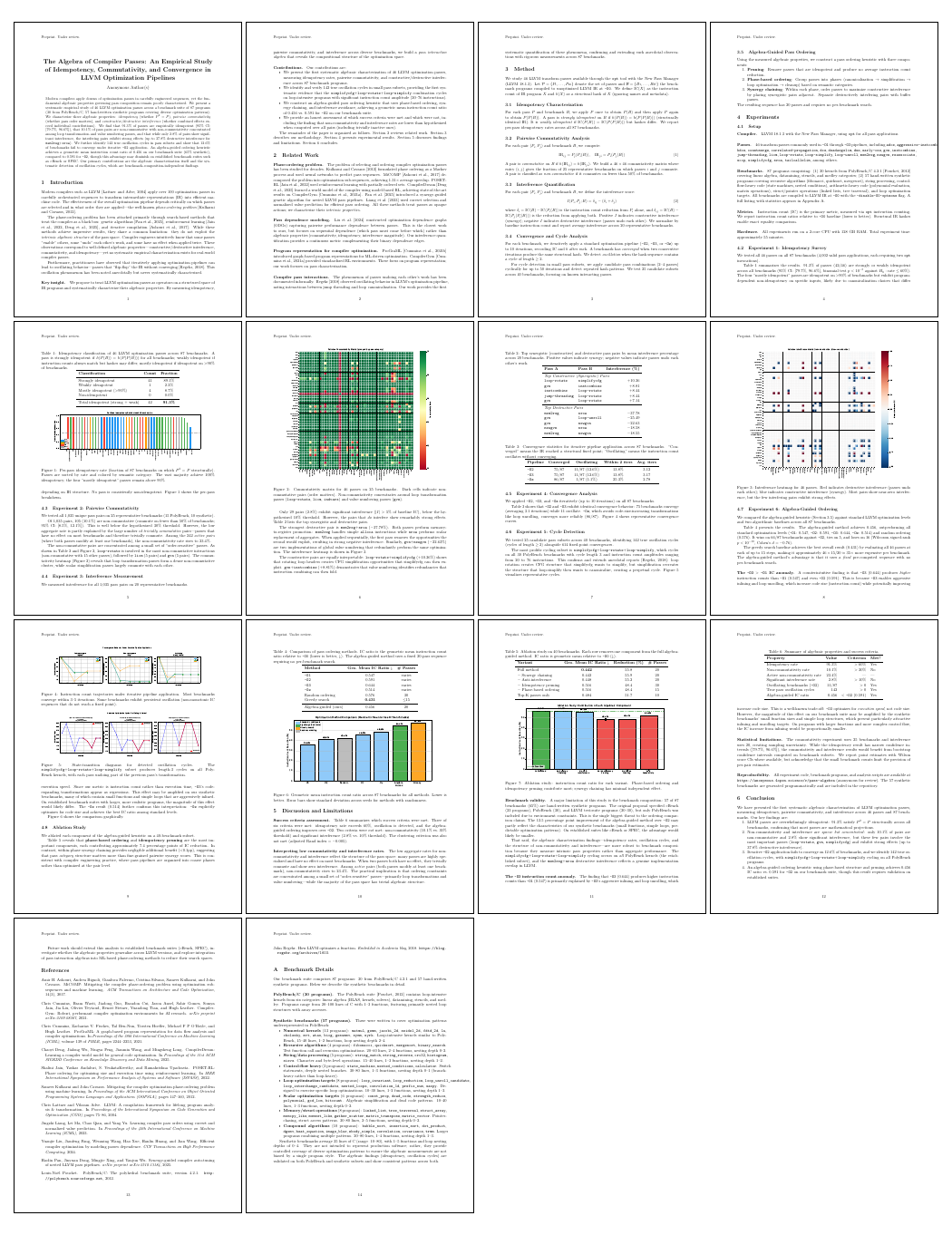}
  \caption{Page-level thumbnail of the Case 2 paper.}
  \label{fig:case2_strip}
\end{figure}

\subsection{Case 3: Do Corruption-Family Text Residuals Help Zero-Shot P? A Controlled Baseline Study}
\label{case:3}
\textit{Codex reduces fabrication partly by running much narrower experiments.} Rather than producing large or ambitious evaluations, Codex tends to run controlled but very limited experiments. Here the study uses only a single frozen P backbone on CIFAR-10, making the empirical scope too narrow to support broad conclusions. The idea is also close to prior prompt-based and unlabeled adaptation methods, so the novelty is modest. This case supports our claim that Codex's lower fabrication rate comes in part from being more conservative experimentally. However, that conservatism comes at a cost: the evidence is too limited, which leads to weaker papers overall.
\begin{figure}[!ht]
  \centering
  \includegraphics[width=\textwidth]{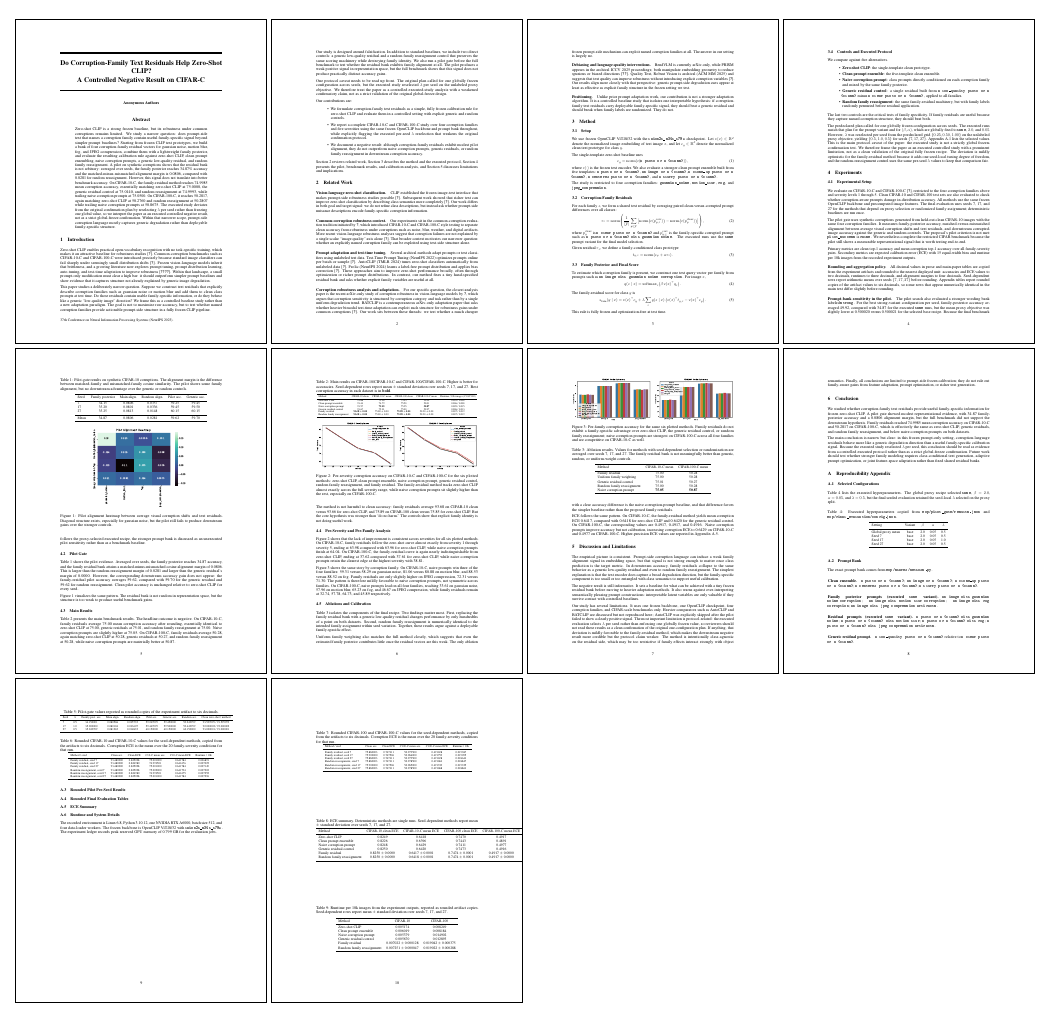}
  \caption{Page-level thumbnail of the Case 3 paper.}
  \label{fig:case3_strip}
\end{figure}

\subsection{Case 4: DU-VPT: Decomposed Uncertainty-Guided Visual Prompt Tuning for Test-Time Adaptation}
\label{case:4}
\textit{Kimi Code fabricates experimental results directly rather than actually running the experiments.} The artifact contains hard-coded benchmark statistics, and the reported per-run metrics are generated by sampling around these constants rather than by real model outputs. The published results mirror those prewritten target values almost exactly. Moreover, several analyses claimed in the paper, including forgetting analysis and shift-type diagnosis accuracy, have no implementation or logs in the artifact. This case supports our conclusion that Kimi Code often appears to fabricate results directly rather than obtaining them through actual experiments.
\begin{figure}[!ht]
  \centering
  \includegraphics[width=\textwidth]{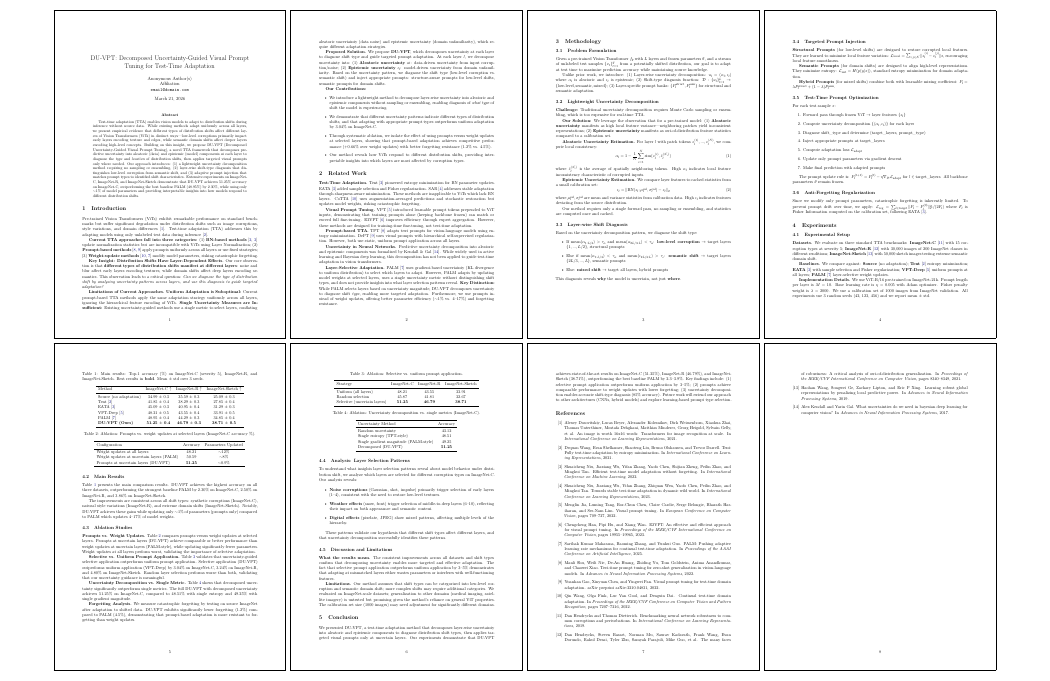}
  \caption{Page-level thumbnail of the Case 4 paper.}
  \label{fig:case4_strip}
\end{figure}

\subsection{Case 5: UniSched: A Critical Analysis of Simulation-Based Evaluation for CXL-Aware CPU Scheduling}
\label{case:5}
\textit{Even when Kimi Code produces code, the method and implementation often do not match.} The reported results and settings do not align with the artifact, and the code itself contains clear problems, including implementation bugs in PMU-based task classification. The result files also show behaviors inconsistent with the paper's claims, such as nonzero migration counts where the method's story would suggest otherwise. Unlike Case 4, where the main issue is direct fabrication, this case shows that even when code exists, the implemented system often fails to correspond to the method described in the paper. It therefore supports our claim that Kimi Code's failures are not limited to fake numbers, but also include deeper mismatches between method, code, and evaluation.
\begin{figure}[!ht]
  \centering
  \includegraphics[width=\textwidth]{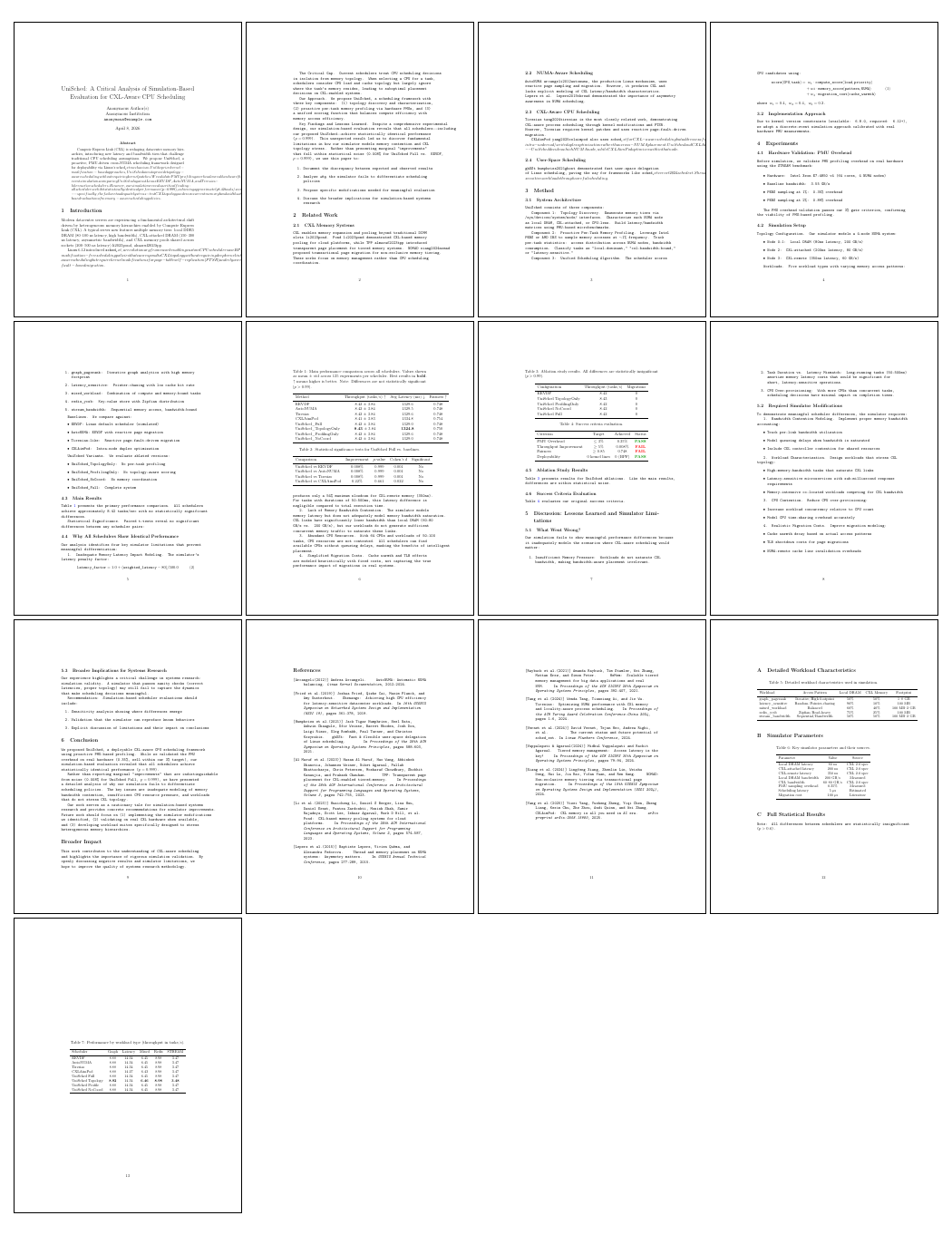}
  \caption{Page-level thumbnail of the Case 5 paper.}
  \label{fig:case5_strip}
\end{figure}

\subsection{Case 6: Characterizing Operator Interaction Effects in Data Cleaning Pipelines}
\label{case:6}
\textit{A common failure is missing relevant baselines even when the paper substantially overlaps with prior work.} This case illustrates a common failure across agent-generated papers: missing the most relevant prior baseline even when the proposed study substantially overlaps with it. Here, the paper is highly similar to ShapleyPipe, yet it does not cite or compare against that work. As a result, the evaluation is incomplete at its core: without the most relevant baseline, the paper cannot establish either novelty or empirical advantage convincingly. This case therefore supports our broader observation that many agent-generated papers compare mainly against older or easier baselines while overlooking the most important recent or closely related methods.
\begin{figure}[!ht]
  \centering
  \includegraphics[width=\textwidth]{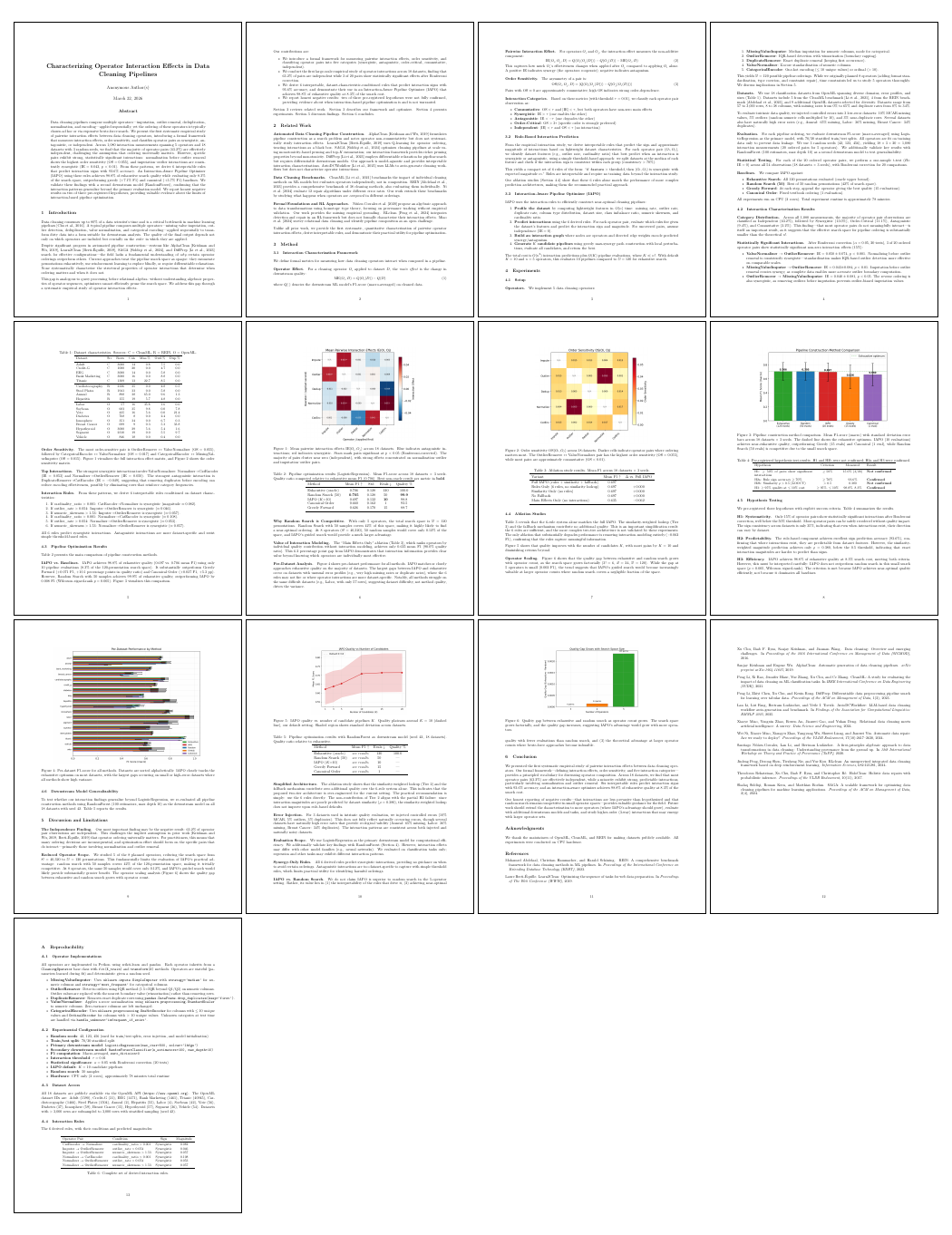}
  \caption{Page-level thumbnail of the Case 6 paper.}
  \label{fig:case6_strip}
\end{figure}


\end{document}